\documentclass[10pt,twocolumn,letterpaper]{article}

\usepackage[pagenumbers]{cvpr} 

\usepackage[accsupp]{axessibility}
\usepackage{graphicx}
\usepackage{amsmath}
\usepackage{amssymb}
\usepackage{booktabs}

\usepackage{multirow}
\usepackage{bm}
\usepackage{pifont}
\usepackage{amsthm}
\usepackage[pagebackref,breaklinks,colorlinks]{hyperref}

\usepackage[capitalize]{cleveref}
\crefname{section}{Sec.}{Secs.}
\Crefname{section}{Section}{Sections}
\Crefname{table}{Table}{Tables}
\crefname{table}{Tab.}{Tabs.}

\def\cvprPaperID{8841} 
\def\confName{CVPR}
\def\confYear{2022}

\begin{document}

\title{Knowledge Distillation with the Reused Teacher Classifier}

\author{    
	Defang Chen\textsuperscript{\rm 1,2,3} \quad
	Jian-Ping Mei\textsuperscript{\rm 4} \quad 
	Hailin Zhang\textsuperscript{\rm 1,2,3} \\
	Can Wang\textsuperscript{\rm 1,2,3}\thanks{Corresponding author} \quad
	Yan Feng\textsuperscript{\rm 1,2,3} \quad
	Chun Chen\textsuperscript{\rm 1,2,3}\\ 
    \textsuperscript{\rm 1}Zhejiang University \quad
    \textsuperscript{\rm 2}Shanghai Institute for Advanced Study of Zhejiang University \\
    \textsuperscript{\rm 3}ZJU-Bangsun Joint Research Center \quad
   \textsuperscript{\rm 4}Zhejiang University of Technology\\ 
{\tt\small defchern@zju.edu.cn, jpmei@zjut.edu.cn, \{zzzhl, wcan, fengyan, chenc\}@zju.edu.cn}
}
\maketitle

\begin{abstract}
Knowledge distillation aims to compress a powerful yet cumbersome teacher model into a lightweight student model without much sacrifice of performance. For this purpose, various approaches have been proposed over the past few years, generally with elaborately designed knowledge representations, which in turn increase the difficulty of model development and interpretation. In contrast, we empirically show that a \textbf{simple knowledge distillation} technique is enough to significantly narrow down the teacher-student performance gap. We directly reuse the discriminative classifier from the pre-trained teacher model for student inference and train a student encoder through feature alignment with a single $\ell_2$ loss. In this way, the student model is able to achieve exactly the same performance as the teacher model provided that their extracted features are perfectly aligned. An additional projector is developed to help the student encoder match with the teacher classifier, which renders our technique applicable to various teacher and student architectures. Extensive experiments demonstrate that our technique achieves state-of-the-art results at the modest cost of compression ratio due to the added projector. 
\end{abstract}

\section{Introduction}

Given a powerful teacher model with large numbers of parameters, the goal of knowledge distillation (KD) is to help another less-parameterized student model gain a similar generalization ability as the larger teacher model \cite{bucilua2006model,hinton2015distilling}. A straightforward way to achieve this goal is by aligning their logits or class predictions given the same inputs \cite{ba2014deep,hinton2015distilling}. 
Due to its conceptual simplicity and practical effectiveness, 
KD technique has achieved great success in a variety of applications, such as object detection \cite{chen2017detection}, semantic segmentation \cite{liu2019structured} and the training of transformers \cite{touvron2021transformers}.
\begin{figure}
	\centering
	\includegraphics[width=0.88\columnwidth]{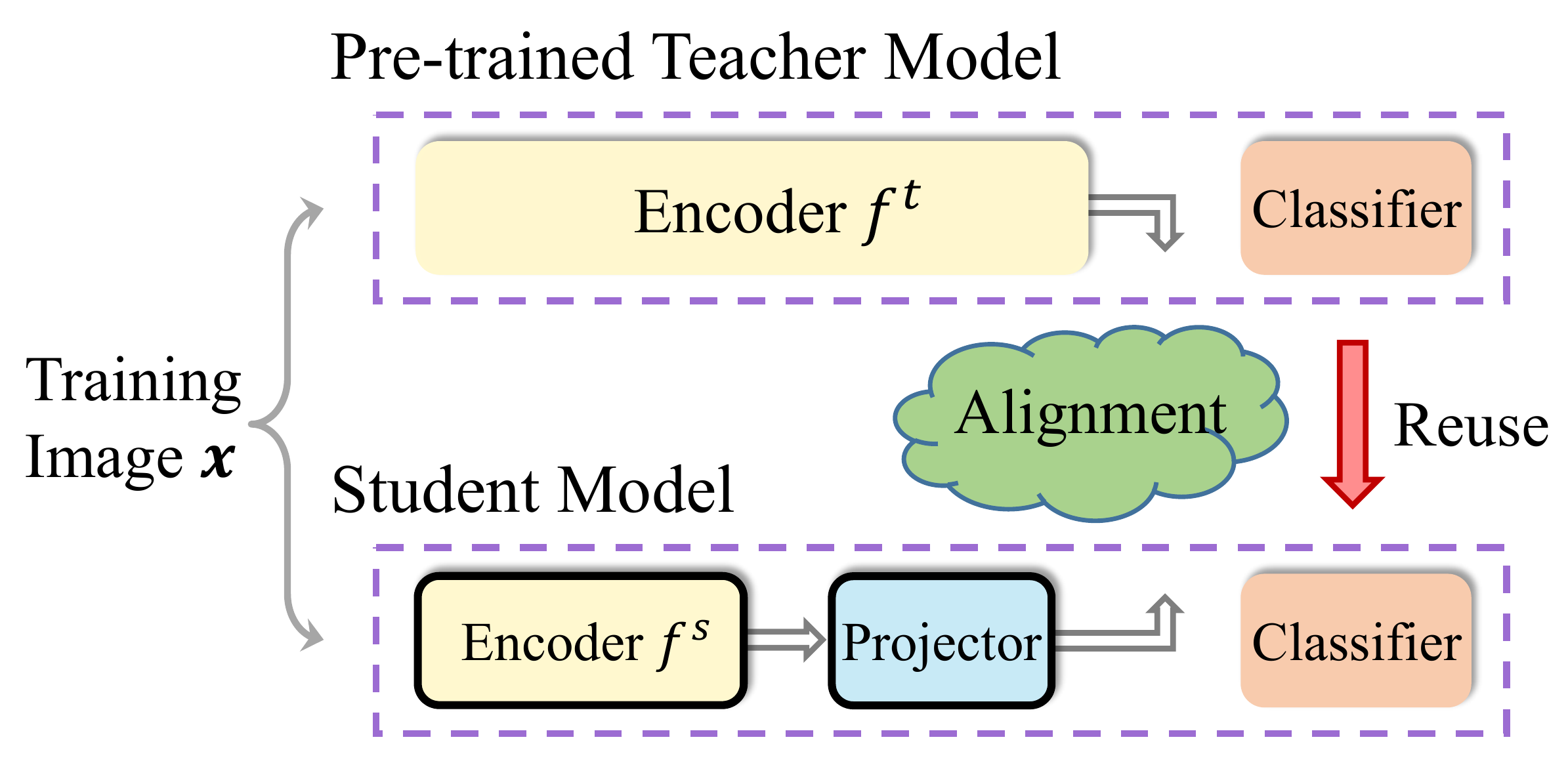}
	\caption{An overview of our proposed SimKD. A simple $\ell_2$ loss is adopted for feature alignment in the preceding layer of the final classifier. Only the student feature encoder and dimension projector are updated during training (boxes with the black border). The pre-trained teacher classifier is reused for student inference.}
	\label{fig:model}
\end{figure}

One limitation of the vanilla KD is that the performance gap between the original teacher model and the distilled student model is still significant. To overcome this drawback, a bunch of approaches have been proposed in the last few years \cite{gou2021survey,wang2021survey}. Most of them benefit from exploiting additional supervision from the pre-trained teacher model, especially the intermediate layers \cite{romero2015fitnets,zagoruyko2017paying,ahn2019variational,tung2019similarity,tian2020contrastive,chen2021cross,yang2021knowledge}. Besides aligning the plain intermediate features \cite{romero2015fitnets,chen2021cross,yang2021knowledge}, the existing efforts are typically based on elaborately designed knowledge representations, such as mimicking spatial attention maps \cite{zagoruyko2017paying}, pairwise similarity patterns \cite{passalis2018learning,park2019relational,tung2019similarity} or maximizing the mutual information between teacher and student features \cite{ahn2019variational,tian2020contrastive,zhou2021holistic}. 
Although we indeed see constant improvements of these works in student performance, \textit{neither} effective representations \textit{nor} well-optimized hyper-parameters ensuring their success are easily achievable in practice. Furthermore, the diversity of transferred knowledge hinders the emergence of a unified and clear interpretation of the final improvement in student performance.

In this paper, we present a simple knowledge distillation technique and demonstrate that it can significantly bridge the performance gap between teacher and student models with no need for elaborate knowledge representations. Our proposed ``SimKD'' technique is illustrated in Figure~\ref{fig:model}. We argue that the powerful class prediction ability of a teacher model is credited to not only those expressive features but just as importantly, a discriminative classifier. Based on this argument, which is empirically supported later on, we train a student model through feature alignment in the preceding layer of the classifier and directly copy the teacher classifier for student inference. In this way, if we could perfectly align the student features with those of the teacher model, their performance gap will just disappear. That is to say, the feature alignment error alone accounts for the accuracy of student inference, which makes our knowledge transfer more comprehensible. According to our experimental results, a single $\ell_2$ loss for feature alignment already works surprisingly well. Such a simple loss saves us from carefully tuning hyper-parameters as previous works do in order to balance the effect of multiple losses \cite{hinton2015distilling,romero2015fitnets,zagoruyko2017paying,ahn2019variational,tung2019similarity,tian2020contrastive,chen2021cross,yang2021knowledge}.

As the dimensions of extracted features from teacher and student models usually differ from each other, a projector is thus added after the student feature encoder to remedy this dimension mismatch. This projector generally incurs a less than 3\% cost to the pruning ratio in teacher-to-student compression, but it makes our technique applicable to arbitrary model architectures. The pruning ratio could be even enlarged in a few cases where the parameter number of the added projector plus the reused teacher classifier is less than that of the original student classifier (see Figure~\ref{fig:count}). We conduct extensive experiments on standard benchmark datasets and observe that our SimKD consistently outperforms all compared state-of-the-art approaches with a variety of teacher-student architecture combinations. We also show that our simple technique generalizes well in different scenarios such as multi-teacher knowledge distillation and data-free knowledge distillation.

\section{Related Work}

Knowledge distillation (KD) is a technique to compress the knowledge from a powerful teacher model, such as an ensemble of multiple deep neural networks, into a smaller student model \cite{bucilua2006model,hinton2015distilling,gou2021survey,wang2021survey}. The transferred knowledge is initially regarded as the conditional distribution of outputs given input samples \cite{hinton2015distilling}. From this viewpoint, the predictions, or soft targets, from the pre-trained teacher model play a major role in the improvement of student performance. A common belief behind the success of this technique is that those teacher-learned soft targets can capture the relationships among different categories and serve as an effective regularization during student training \cite{ba2014deep,hinton2015distilling,chen2020online,yuan2020revisiting}.


In order to make KD more practical for model compression, we need to further resist the performance degradation in teacher-to-student compression \cite{gou2021survey,wang2021survey}. Leveraging more information from the pre-trained teacher model especially the intermediate layers is a general solution towards this problem. A bunch of such works have sprung up seeking for better student performance in the last few years, collectively known as \textit{feature distillation}. They mostly propose diverse representations to capture appropriate transferred knowledge, such as the crude intermediate feature maps \cite{romero2015fitnets} or their transformations \cite{zagoruyko2017paying,heo2019knowledge,ahn2019variational}, sample relations encoded by the pairwise similarity matrices \cite{park2019relational,tung2019similarity,passalis2018learning} or modeled by contrastive learning \cite{tian2020contrastive,xu2020knowledge,zhou2021holistic}. More recently, a few works turn to designing cross-layer associations to make full use of those intermediate features of the teacher model \cite{chen2021cross,chen2021review}. With the help of aforementioned knowledge representations or reformed transfer strategies, the student model will be trained with gradient information coming from not only the final layer, i.e., the classifier, but also from those early layers. However, additional hyper-parameters need careful tuning in these methods to balance the effect of different losses and it is still unclear how the newly introduced supervisory signal would exert positive influence on the final performance of student models.

To some extent, our key idea of reusing the teacher classifier is related to the previous studies on hypothesis transfer learning (HTL) \cite{redko2020survey}. HTL aims to utilize the learned source domain classifier to help the training of the target domain classifier, on the condition that only a small amount of labeled target dataset and no source dataset are accessible \cite{kuzborskij2013stability,du2017hypothesis,kuzborskij2017fast}. A recent work further gets rid of the requirement of labeling target dataset and extends the vanilla HTL to the unsupervised domain adaptation setting by resorting to a pseudo-labeling strategy \cite{liang2020do}. Different from this one, our goal is to reduce the teacher-student performance gap on the same dataset, rather than adapting the pre-trained model to achieve good performance on another dataset with a different distribution. In addition, our SimKD is much simpler than this work and still achieves surprisingly good results in the standard KD setting. 
\section{Method}
\begin{figure*}
	\centering
	\includegraphics[width=0.99\linewidth]{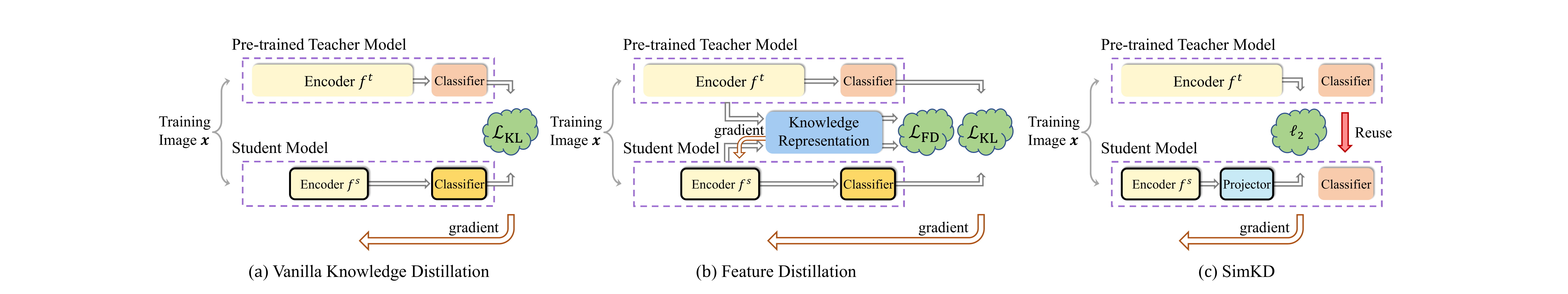}
	\caption{Comparison of three kinds of knowledge distillation techniques. The main differences lie in how the gradient is formalized and where the gradient flow starts. (a) Vanilla KD calculates the gradient in class predictions and relies on this gradient to update the whole student model. (b) Feature distillation gathers more gradient information from the intermediate layers through various knowledge representations. Additional hyper-parameters need to be carefully turned for maximum performance. (c) Our SimKD calculates $\ell_2$ loss in the preceding layer of the classifier and back propagates this gradient solely to update the student feature encoder and dimension projector. 
	The cross entropy losses between predictions and ground-truth labels of the compared approaches in (a) and (b) are omitted for simplicity.
	}
	\label{fig:model-cmp}
\end{figure*}

\subsection{Vanilla Knowledge Distillation}
Generally, the popular deep neural networks designed for image classification tasks in the current era can be regarded as the stack of a \textit{feature encoder} with multiple non-linear layers, together with a \textit{classifier} that usually contains a single fully-connected layer with softmax activation function \cite{he2016deep,huang2017densely,sandler2018mobile,zhang2018shufflenet,ma2018shuffle}. Both two components will be trained end-to-end with the back-propagation algorithm. The symbolic description is presented as follows.

Given a training sample $\boldsymbol{x}$ with one-hot label $\boldsymbol{y}$ from a $K$-category classification dataset, we denote the encoded feature in the penultimate layer of the student model as $\boldsymbol{f}^{s}=\mathcal{F}^{s}(\boldsymbol{x};\boldsymbol{\theta}^s)\in \mathbb{R}^{C_{s}}$. This feature is subsequently passed into the classifier with weight $\boldsymbol{W}^{s}\in \mathbb{R}^{K\times C_{s}}$ to obtain the logits $\boldsymbol{g}^{s}=\boldsymbol{W}^{s}\boldsymbol{f}^{s}\in \mathbb{R}^{K}$ as well as the class prediction $\boldsymbol{p}^{s}=\sigma(\boldsymbol{g}^{s}/T)\in \mathbb{R}^{K}$ with a softmax activation function $\sigma(\cdot)$ and the temperature $T$
\begin{equation}
	p^{s}_{i}=\frac{\exp\left(g^{s}_{i}/T\right)}{\sum_{j=1}^{K}\exp \left(g^{s}_{j}/T\right)},
\end{equation}
where $p^{s}_{i}/g^{s}_{i}$ denotes the $i$-th element of corresponding vectors and $T$ is a hyper-parameter for softening effect\footnote{We only present notations for the student model in this paragraph, but similar notations also hold for the teacher model.}.

Vanilla knowledge distillation consists of two losses \cite{hinton2015distilling}: one is the conventional cross entropy loss and another is the alignment loss in prediction pairs between $\boldsymbol{p}^{s}$ and soft targets $\boldsymbol{p}^{t}$ with Kullback-Leibler divergence \cite{kullback1951information} %
\begin{equation}
	\label{eq:kd}
	\begin{aligned} 			
		\mathcal{L}_{\mathrm{KD}}=\underbrace{\mathcal{L}_{\mathrm{CE}}(\boldsymbol{y}, \boldsymbol{p}^s)}_{T=1}+\underbrace{T^2\mathcal{L}_{\mathrm{KL}}(\boldsymbol{p}^{t}, \boldsymbol{p}^{s})}_{T>1}.
	\end{aligned}
\end{equation}
Compared to the cross entropy loss, the introduced prediction alignment loss gives extra information on incorrect classes to facilitate the student training \cite{hinton2015distilling,furlanello2018born}. Since probabilities assigned to those incorrect classes tend to be rather small after softmax transformation, the temperature $T$ in this term needs raising to produce softer distributions for conveying more information \cite{hinton2015distilling}. 

\subsection{Simple Knowledge Distillation}
In recent years, various feature distillation approaches have been proposed. These works mainly collect and transmit extra gradient information from intermediate teacher-student layer pairs to train the student feature encoder better (Figure~\ref{fig:model-cmp}b). However, their success heavily depends on those particularly-designed knowledge representations to entail proper inductive bias \cite{bengio13representation,chen2021cross}, and carefully chosen hyper-parameters to balance the effect of different losses. Both are labor-intensive and time-consuming. It is also difficult to conclude the actual role that a certain type of representation plays in the student training.

In contrast, we propose a simple knowledge distillation technique named as SimKD, which breaks away from these stringent demands while still achieving state-of-the-art results on extensive experiments. As shown in Figure~\ref{fig:model-cmp}c, a key ingredient of SimKD is the ``\textit{classifier-reusing}" operation, \ie, we directly borrow the pre-trained teacher classifier for student inference rather than training a new one. This eliminates the need of label information to calculate the cross entropy loss and makes the feature alignment loss become the only source for generating gradient.

Overall, we argue discriminative information contained in the teacher classifier matters, but has been largely overlooked in the literature of KD. We then provide a plausible explanation for its important role. Consider a situation where one model is requested to handle several tasks with different data distributions, a basic practice is to freeze or share some shallow layers as the feature extractor across different tasks while fine-tuning the last layer to learn \textit{task-specific} information \cite{caruana1997multitask,girshick2014rich,donahue2014decaf,li2018learning}. In this one-model multiple-task setting, existing works hold the opinion that \textit{task-invariant} information could be shared while task-specific information needs to be independently identified, generally by the final classifier. As for KD where teacher and student models with different capabilities are trained on the same dataset, analogously, we could reasonably believe that there is some \textit{capability-invariant} information in the data being easily gained across different models while the powerful teacher model may contain extra essential \textit{capability-specific} information that is hard for a simpler student model to acquire. 
Furthermore, we hypothesize that most capability-specific information is contained in deep layers and expect that reusing these layers, even only the final classifier will be helpful for the student training.

\begin{figure}
	\centering
	\begin{subfigure}{0.48\columnwidth}
		\includegraphics[width=0.9\columnwidth]{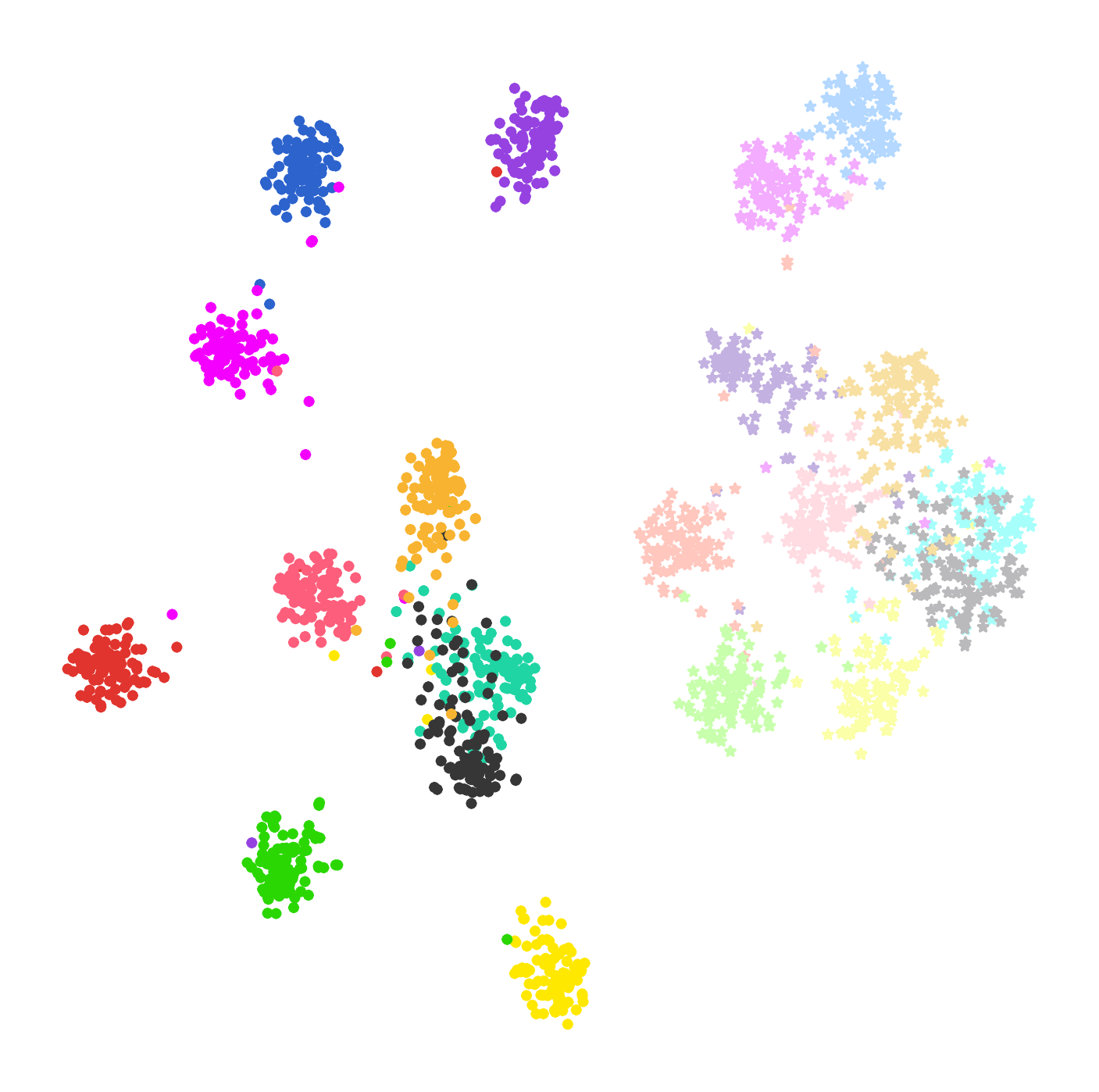}
		\caption{Vanilla KD \cite{hinton2015distilling}.}
		\label{fig:vis-a}
	\end{subfigure}
	\hfill
	\begin{subfigure}{0.48\columnwidth}
		\includegraphics[width=0.9\columnwidth]{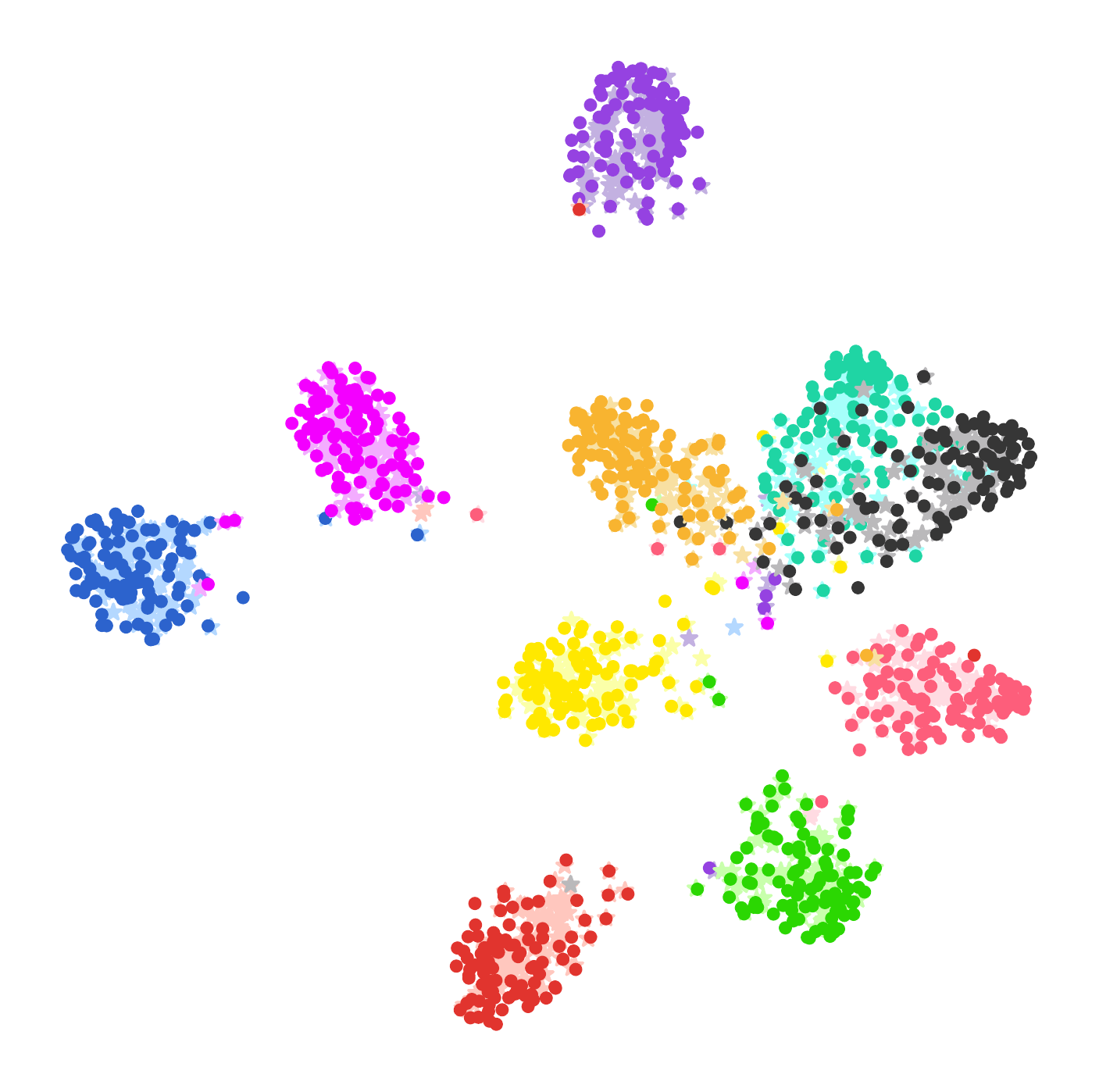}
		\caption{Our SimKD.}
		\label{fig:vis-b}
	\end{subfigure}
	\caption{Visualization results of test images from CIFAR-100 with t-SNE \cite{vandermaaten2008visual}. We randomly sample 10 out of 100 classes. Features extracted by the teacher and student models are depicted with dark and light colors, respectively, and they are almost indistinguishable in our SimKD. Best viewed in color.}
	\label{fig:vis}
\end{figure}

Based on this hypothesis, which is supported later by empirical evidences from various aspects, we furnish the student model with the teacher classifier for inference and force their extracted features to be matched with the following $\ell_2$ loss function
\begin{equation}
	\label{eq:simKD}
	\begin{aligned} 				\mathcal{L}_{\mathrm{SimKD}}=\|\boldsymbol{f}^{t}-\mathcal{P}(\boldsymbol{f}^{s})\|^2_2,
	\end{aligned}
\end{equation}
where a projector $\mathcal{P}(\cdot)$ is designed to match the feature dimensions at a relatively small cost while being effective enough to ensure accurate alignment. In effect, this simple loss has already been exploited before \cite{romero2015fitnets,yang2021knowledge}, but we are actually attempting to reveal the potential value of reusing the teacher classifier rather than developing a sophisticated loss function for feature alignment.
As shown in Figure~\ref{fig:vis}, the extracted features from the pre-trained teacher model (dark colors) and the distilled student model in our SimKD (light colors) are compactly clustered within the same class and distinctly separated across different classes, which ensures the student features to be correctly classified latter with the reused teacher classifier.

Somewhat surprisingly, the performance degradation in teacher-to-student compression will be greatly alleviated by this simple technique. Along with high inference accuracy, the simplicity of this single-loss formulation provides our SimKD with good interpretability. Note that the reused part from a pre-trained teacher model is allowed to incorporate more layers but not just limited to the final classifier. Usually, reusing more layers leads to higher student accuracy, but will bring about the burden increase on the inference. 

\begin{table*}[htbp]
	\centering
	\resizebox{0.95\textwidth}{!}{
		\begin{tabular}{c|ccccccc}
			\toprule
			\multirow{2}{*}{Student} &  WRN-40-1 & ResNet-8x4  & ResNet-110 & ResNet-116 & VGG-8 & ResNet-8x4 & ShuffleNetV2 \\
			& 71.92 $\pm$ 0.17 & 73.09 $\pm$ 0.30 & 74.37 $\pm$ 0.17 & 74.46 $\pm$ 0.09 & 70.46 $\pm$ 0.29 & 73.09 $\pm$ 0.30 & 72.60 $\pm$ 0.12  \\
			\midrule
			KD \cite{hinton2015distilling} & 74.12 $\pm$ 0.29 & 74.42 $\pm$ 0.05 & 76.25 $\pm$ 0.34 & 76.14 $\pm$ 0.32 & 72.73 $\pm$ 0.15 & 75.28 $\pm$ 0.18 & 75.60 $\pm$ 0.21  \\
			FitNet \cite{romero2015fitnets} & 74.17 $\pm$ 0.22 & 74.32 $\pm$ 0.08 & 76.08 $\pm$ 0.13 & 76.20 $\pm$ 0.17 & 72.91 $\pm$ 0.18 & 75.02 $\pm$ 0.31 & 75.82 $\pm$ 0.22  \\			
			AT \cite{zagoruyko2017paying} & 74.67 $\pm$ 0.18 & 75.07 $\pm$ 0.03  & 76.67 $\pm$ 0.28 & 76.84 $\pm$ 0.25 & 71.90 $\pm$ 0.13 & 75.74 $\pm$ 0.09 & 75.41 $\pm$ 0.10  \\
			SP \cite{tung2019similarity} & 73.90 $\pm$ 0.17 & 74.29 $\pm$ 0.07  & 76.43 $\pm$ 0.39 & 75.99 $\pm$ 0.26 & 73.12 $\pm$ 0.10 & 74.84 $\pm$ 0.08 & 75.77 $\pm$ 0.08  \\
			VID \cite{ahn2019variational} & 74.59 $\pm$ 0.17 & 74.55 $\pm$ 0.10 & 76.17 $\pm$ 0.22 & 76.53 $\pm$ 0.24 & 73.19 $\pm$ 0.23 & 75.56 $\pm$ 0.13 & 75.22 $\pm$ 0.07  \\
			CRD \cite{tian2020contrastive} & 74.80 $\pm$ 0.33  & 75.59 $\pm$ 0.07 & 76.86 $\pm$ 0.09 & 76.83 $\pm$ 0.13 & 73.54 $\pm$ 0.19 & 75.78 $\pm$ 0.27 & 77.04 $\pm$ 0.61  \\
			SRRL \cite{yang2021knowledge} & 74.64 $\pm$ 0.14 & 75.39 $\pm$ 0.34  & 76.75 $\pm$ 0.14 & 77.19 $\pm$ 0.09 & 73.23 $\pm$ 0.16 & 76.12 $\pm$ 0.18 & 76.19 $\pm$ 0.35 \\
			SemCKD \cite{chen2021cross} & 74.41 $\pm$ 0.16 & 76.23 $\pm$ 0.04  & 76.62 $\pm$ 0.14 & 76.69 $\pm$ 0.48 & 75.27 $\pm$ 0.13 & 75.85 $\pm$ 0.16 & 77.62 $\pm$ 0.32  \\
			\midrule
			SimKD & \textbf{75.56 $\pm$ 0.27} & \textbf{78.08 $\pm$ 0.15}  & \textbf{77.82 $\pm$ 0.15} & \textbf{77.90 $\pm$ 0.11} & \textbf{75.76 $\pm$ 0.12} & \textbf{76.75 $\pm$ 0.23} & \textbf{78.39 $\pm$ 0.27}  \\
			\midrule
			\multirow{2}{*}{Teacher} & WRN-40-2 & ResNet-32x4 & ResNet-110x2 & ResNet-110x2 & ResNet-32x4 & WRN-40-2 & ResNet-32x4  \\
			& 76.31 & 79.42 & 78.18 & 78.18 & 79.42 & 76.31 & 79.42  \\
			\bottomrule
		\end{tabular}
	}
	\caption{Top-1 test accuracy (\%) of various knowledge distillation approaches on CIFAR-100.}
	\label{Table:CIFAR-100}
\end{table*} 
\begin{table*}[t]
	\centering
	\resizebox{0.95\textwidth}{!}{
		\begin{tabular}{c|ccccccc}
			\toprule
			\multirow{2}{*}{Student} & ShuffleNetV1 & WRN-16-2 & ShuffleNetV2 & MobileNetV2 & \small MobileNetV2x2 & WRN-40-2 & \small ShuffleNetV2x1.5 \\
			& 71.36 $\pm$ 0.25 & 73.51 $\pm$ 0.32 & 72.60 $\pm$ 0.12 & 65.43 $\pm$ 0.29 & 69.06 $\pm$ 0.10 & 76.35 $\pm$ 0.18  & 74.15 $\pm$ 0.22 \\
			\midrule
			KD \cite{hinton2015distilling} & 74.30 $\pm$ 0.16 & 74.90 $\pm$ 0.29 & 76.05 $\pm$ 0.34 & 69.07 $\pm$ 0.47 & 72.43 $\pm$ 0.32 & 77.70 $\pm$ 0.13  & 76.82 $\pm$ 0.23 \\
			FitNet \cite{romero2015fitnets} & 74.52 $\pm$ 0.03 & 74.70 $\pm$ 0.35 & 76.02 $\pm$ 0.21 & 68.64 $\pm$ 0.27 & 73.09 $\pm$ 0.46 & 77.69 $\pm$ 0.23  & 77.12 $\pm$ 0.24 \\
			AT \cite{zagoruyko2017paying} & 75.55 $\pm$ 0.19  & 75.38 $\pm$ 0.18 & 76.84 $\pm$ 0.19 & 68.62 $\pm$ 0.31 & 73.08 $\pm$ 0.14 & 78.45 $\pm$ 0.24 & 77.51 $\pm$ 0.31 \\
			SP \cite{tung2019similarity} & 74.69 $\pm$ 0.32 & 75.16 $\pm$ 0.32 & 76.60 $\pm$ 0.22 & 68.73 $\pm$ 0.17 & 72.99 $\pm$ 0.27 & 78.34 $\pm$ 0.08  & 77.18 $\pm$ 0.19 \\
			VID \cite{ahn2019variational} & 74.76 $\pm$ 0.22 & 74.85 $\pm$ 0.35 & 76.44 $\pm$ 0.32 & 68.91 $\pm$ 0.33 & 72.70 $\pm$ 0.22 & 77.96 $\pm$ 0.33  & 77.11 $\pm$ 0.35 \\
			CRD \cite{tian2020contrastive} & 75.34 $\pm$ 0.24 & 75.65 $\pm$ 0.08 & 76.67 $\pm$ 0.27 & 70.28 $\pm$ 0.24 & 73.67 $\pm$ 0.26 & 78.15 $\pm$ 0.14  & 77.66 $\pm$ 0.22 \\
			SRRL \cite{yang2021knowledge} & 75.18 $\pm$ 0.39 & 75.46 $\pm$ 0.13 & 76.71 $\pm$ 0.27 & 69.34 $\pm$ 0.16 & 73.48 $\pm$ 0.36 & 78.39 $\pm$ 0.19 & 77.55 
			$\pm$ 0.26\\
			SemCKD \cite{chen2021cross} & 76.31 $\pm$ 0.20 & 75.65 $\pm$ 0.23 & 77.67 $\pm$ 0.30 & 69.88 $\pm$ 0.30 & 73.98 $\pm$ 0.32 & 78.74 $\pm$ 0.17 & 79.13 $\pm$ 0.41 \\
			\midrule
			SimKD & \textbf{77.18 $\pm$ 0.26} & \textbf{77.17 $\pm$ 0.32} & \textbf{78.25 $\pm$ 0.24} & \textbf{70.71 $\pm$ 0.41} & \textbf{75.43 $\pm$ 0.26} & \textbf{79.29 $\pm$ 0.11} & \textbf{79.54 $\pm$ 0.26}\\
			\midrule
			\multirow{2}{*}{Teacher} & ResNet-32x4 & ResNet-32x4 & ResNet-110x2 & WRN-40-2 & ResNet-32x4 & ResNet-32x4 & ResNet-32x4 \\
			& 79.42 & 79.42 & 78.18 & 76.31 & 79.42 & 79.42 & 79.42 \\
			\bottomrule
		\end{tabular}
	}
	\caption{Top-1 test accuracy (\%) of various knowledge distillation approaches on CIFAR-100.}
	\label{Table:CIFAR-100-2}
\end{table*}

\begin{table*}
	\centering
	\resizebox{0.96\textwidth}{!}{
		\begin{tabular}{c|ccccccccc|c}
		\toprule
		& Student & KD \cite{hinton2015distilling} & AT \cite{zagoruyko2017paying} & SP \cite{tung2019similarity} & VID \cite{ahn2019variational} & CRD \cite{tian2020contrastive} & SRRL \cite{yang2021knowledge} & SemCKD \cite{chen2021cross} & SimKD & Teacher \\
		\midrule
		1/4 Epoch & 49.34 & 52.75 & 52.85 & 53.57 & 53.22 & 55.44 & 55.14 & 53.14 & \textbf{61.73} & 54.50 \\
		1/2 Epoch & 64.98 & 66.69 & 66.69 & 66.36 & 66.64 & 67.25 & 67.36 & 66.89 & \textbf{69.26} & 70.55 \\
		Full Epoch & 70.58 & 71.29 & 71.18 & 71.08 & 71.11 & 71.25 & 71.46 & 71.41 & \textbf{71.66} & 76.26 \\		
		\bottomrule
	\end{tabular}
	}
	\caption{Top-1 test accuracy (\%) comparison on ImageNet for different training epochs. We adopt ResNet-18 as the student model.}
	\label{Table:ImageNet}
\end{table*}   

\section{Experiments}
In this section, we conduct extensive experiments to demonstrate the effectiveness of our proposed SimKD. We first compare it with several representative state-of-the-art approaches on standard benchmark datasets. Some empirical evidences are then given to show the superiority of our ``classifier-reusing'' operation on the student performance improvement. Although an additional projector is required for our student inference, experiments show that its effect to the pruning ratio could be controlled in an acceptable level. Finally, we employ our technique to the multi-teacher and data-free knowledge distillation settings.

\textbf{Datasets and baselines.}
Two benchmark image classification datasets including CIFAR-100 \cite{krizhevsky2009learning} and ImageNet \cite{russakovsky2015ImageNet} are adopted for a series of experiments. We use the standard data augmentation and normalize all images by channel means and standard deviations as \cite{he2016deep,zagoruyko2016wide,huang2017densely}. Besides the vanilla KD \cite{hinton2015distilling}, various approaches are reproduced for comparison, including 
FitNet \cite{romero2015fitnets}, AT \cite{zagoruyko2017paying}, SP \cite{tung2019similarity}, VID \cite{ahn2019variational}, CRD \cite{tian2020contrastive}, SRRL \cite{yang2021knowledge} and SemCKD \cite{chen2021cross}. All compared approaches except KD itself are implemented incorporating the vanilla KD loss, \ie, \cref{eq:kd}. 

\textbf{Training details.}
We follow the training procedure of previous works \cite{tian2020contrastive,chen2021cross,yang2021knowledge} and report the performance of all competitors on our randomly associated teacher-student combinations. 
Specifically, we adopt SGD optimizer with 0.9 Nesterov momentum for all datasets. For CIFAR-100, the total training epoch is set to 240 and the learning rate is divided by 10 at 150th, 180th and 210th epochs. The initial learning rate is set to 0.01 for MobileNet/ShuffleNet-series architectures and 0.05 for other architectures. The mini-batch size is set to 64 and the weight decay is set to $5\times10^{-4}$. For ImageNet, the initial learning rate is set to 0.1 and then divided by 10 at 30th, 60th, 90th of the total 120 training epochs. The mini-batch size is set to 256 and the weight decay is set to $1\times 10^{-4}$. 
All results are reported in means (standard deviations) over 4 trials, except for the results on ImageNet are reported in a single trial. The temperature $T$ in the KD loss is set to $4$ throughout this paper. \textit{More detailed descriptions for reproducibility as well as more results are included in the technical appendix.} 

\subsection{Comparison of Test Accuracy}
Table~\ref{Table:CIFAR-100} to~\ref{Table:ImageNet} present a comprehensive performance comparison of various approaches based on \textit{fifteen} network combinations, where the teacher and student models are instantiated with similar or completely different architectures. 


From the test accuracy comparison in Table~\ref{Table:CIFAR-100} and~\ref{Table:CIFAR-100-2}, we can see that SimKD consistently outperforms all competitors on CIFAR-100 and the improvements are quite significant in some cases. For example, as for the ``ResNet-8x4 \& ResNet-32x4'' combination, SimKD achieves 3.66\% absolute accuracy improvement while the best competitor only achieves 1.81\% absolute improvement on the basis of the vanilla KD. 
Moreover, as shown in the fourth and fifth columns of Table~\ref{Table:CIFAR-100}, given the same teacher model ``ResNet-110x2'', SimKD could train a lightweight student model ``ResNet-110'' with a projector containing 0.05M additional parameters to surpass all the competitors by a considerable margin even when they are employed on a ``ResNet-116'' containing about more 0.10M parameters than ``ResNet-110''. 
Test accuracy for different training epochs in Table~\ref{Table:ImageNet} show that SimKD achieves faster convergence in training. 

We also find that the student model trained with SimKD yields higher accuracy than its teacher model in the case of ``ResNet-8x4 \& WRN-40-2'' and ``ShuffleNetV2 \& ResNet-110x2'' combinations, which seems a bit confusing since even zero feature alignment loss only guarantees their accuracies to be exactly the same. A possible explanation from self-distillation is that the feature re-representation effect in Equation (\ref{eq:simKD}) may help the student model become more robust and thus achieve better results \cite{mobahi2020self,deng2021learning}.

\subsection{Classifier-Reusing Operation Analysis}
The ``\textit{classifier-reusing}'' operation is our recipe for success in above performance comparisons. To better understand its crucial role, we conduct several experiments with two alternative strategies to deal with the student feature encoder and classifier: (1) joint training, (2) sequential training. The performance degradation resulted from these two variants confirms the value of discriminative information in the teacher classifier. Moreover, reusing more deep teacher layers will further improve the student performance.
\begin{figure}[t]
	\centering
	\includegraphics[width=0.96\columnwidth]{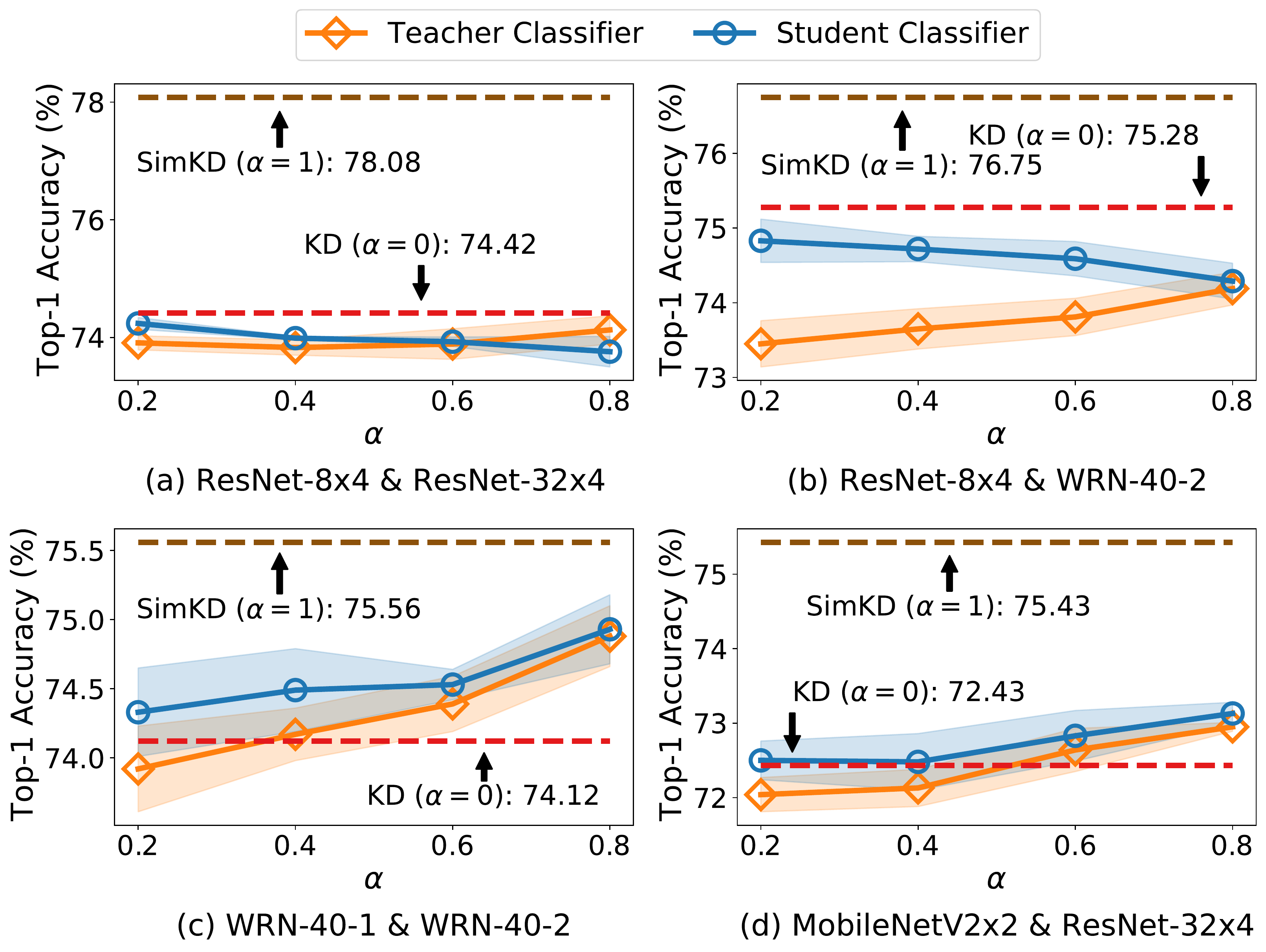}
	\caption{We train the student feature encoder with its associated classifier jointly and then report the test accuracies of student models by using their own classifiers or the reused teacher classifiers.}
	\label{fig:linear}
\end{figure}

\textbf{Joint training.}
As the previous feature distillation approaches do (Figure~\ref{fig:model-cmp}b), we now train the student feature encoder and its associated classifier jointly. The results are obtained by training student models with an extra KD loss 
\begin{equation}
	\begin{aligned} 	
		\mathcal{L}_{\mathrm{Joint}}= (1-\alpha) \mathcal{L}_{\mathrm{KD}} + \alpha \mathcal{L}_{\mathrm{SimKD}},
	\end{aligned}
\end{equation}
where $\alpha$ is a hyper-parameter. 
To thoroughly assess the joint training effect, four different teacher-student combinations together with four uniformly-spaced $\alpha$ values are used. 

As shown in Figure~\ref{fig:linear}, the student performance based on whether its own classifier or the reused teacher classifier becomes far inferior to that of SimKD in all settings, which indicates that discriminative information in the teacher classifier might not be easily transferred into the student model in a joint training way. The substantial accuracy reduction also indicates that the added projector itself and the feature alignment loss do not necessarily improve the final performance, unless we discard joint training and resort to a more effective strategy, \ie, using a single feature alignment loss for training and reusing the teacher classifier for inference. 
Figure~\ref{fig:linear} also shows that in order to surpass the performance of the vanilla KD, this two-loss approach requires a case-by-case hyper-parameter tuning.   

\begin{table}
	\centering
	\resizebox{0.46\textwidth}{!}{
		\begin{tabular}{cccc}
			\toprule
			Student & Sequential  & SimKD & Teacher \\
			\midrule
			\small WRN-40-1 & 74.48 $\pm$ 0.04 & 75.56 $\pm$ 0.27 & \small WRN-40-2  \\
			\small ResNet-8x4 & 51.97 $\pm$ 0.19 & 78.08 $\pm$ 0.15 & \small ResNet-32x4  \\
			\small ResNet-110 & 77.63 $\pm$ 0.05 &  77.82 $\pm$ 0.15 & \small ResNet-110x2  \\
			\small ResNet-116 & 77.75 $\pm$ 0.03 &  77.90 $\pm$ 0.11 & \small ResNet-110x2  \\
			\small VGG-8 & 35.72 $\pm$ 1.33 & 75.76 $\pm$ 0.12 & \small ResNet-32x4  \\
			\small ResNet-8x4 & 45.03 $\pm$ 0.44 & 76.75 $\pm$ 0.23 & \small WRN-40-2  \\
			\small ShuffleNetV2 & 21.56 $\pm$ 0.31 & 78.39 $\pm$ 0.27 & \small ResNet-32x4  \\
			\bottomrule
	\end{tabular}}
	\caption{Training a new student classifier from scratch.}
	\label{Table:sequential}
\end{table}   

\textbf{Sequential training.} The above results show the benefit of disassembling the training of student feature encoder and classifier. Additionally, the ``\textit{classifier-reusing}'' operation carries the implication that a classifier with good discriminative ability is fairly hard to acquire. In this part, we provide evidence for this belief by training a new classifier from scratch rather than reusing the teacher classifier. 

We adopt those teacher-student combinations in Table~\ref{Table:CIFAR-100} as examples for evaluation. After performing feature alignment with Equation~(\ref{eq:simKD}), we fix the student feature encoder, \ie, freeze the extracted features, and train a randomly initialized student classifier (a fully-connected layer with softmax activation) with the regular training procedure. This is exactly same as the linear evaluation protocol used in unsupervised learning evaluation \cite{he2020moco,chen2020simclr,grill2020byon}. 

The results of this sequential training are given in Table~\ref{Table:sequential}. We find that apart from ``WRN-40-1 \& WRN-40-2'' and ``ResNet-110/116 \& ResNet-110x2'', the test accuracies of other student models appear a precipitous drop. Although we have tried tuning the initial learning rate a few times, it only makes a slight difference in performance. Results in Table~\ref{Table:sequential} indicate that even when the extracted features have been aligned, it is still a challenge to train a satisfactory student classifier. Generally, we could achieve better student performance by tuning hyper-parameters in the classifier training step more carefully, but it is a non-trivial task. In contrast, directly reusing the pre-trained teacher classifier already works quite well. \textit{Detailed training procedure and more results are provided in the appendix.}


\begin{figure}[t]
	\begin{minipage}{0.59\linewidth}
		\centering
		\includegraphics[width=0.98\linewidth]{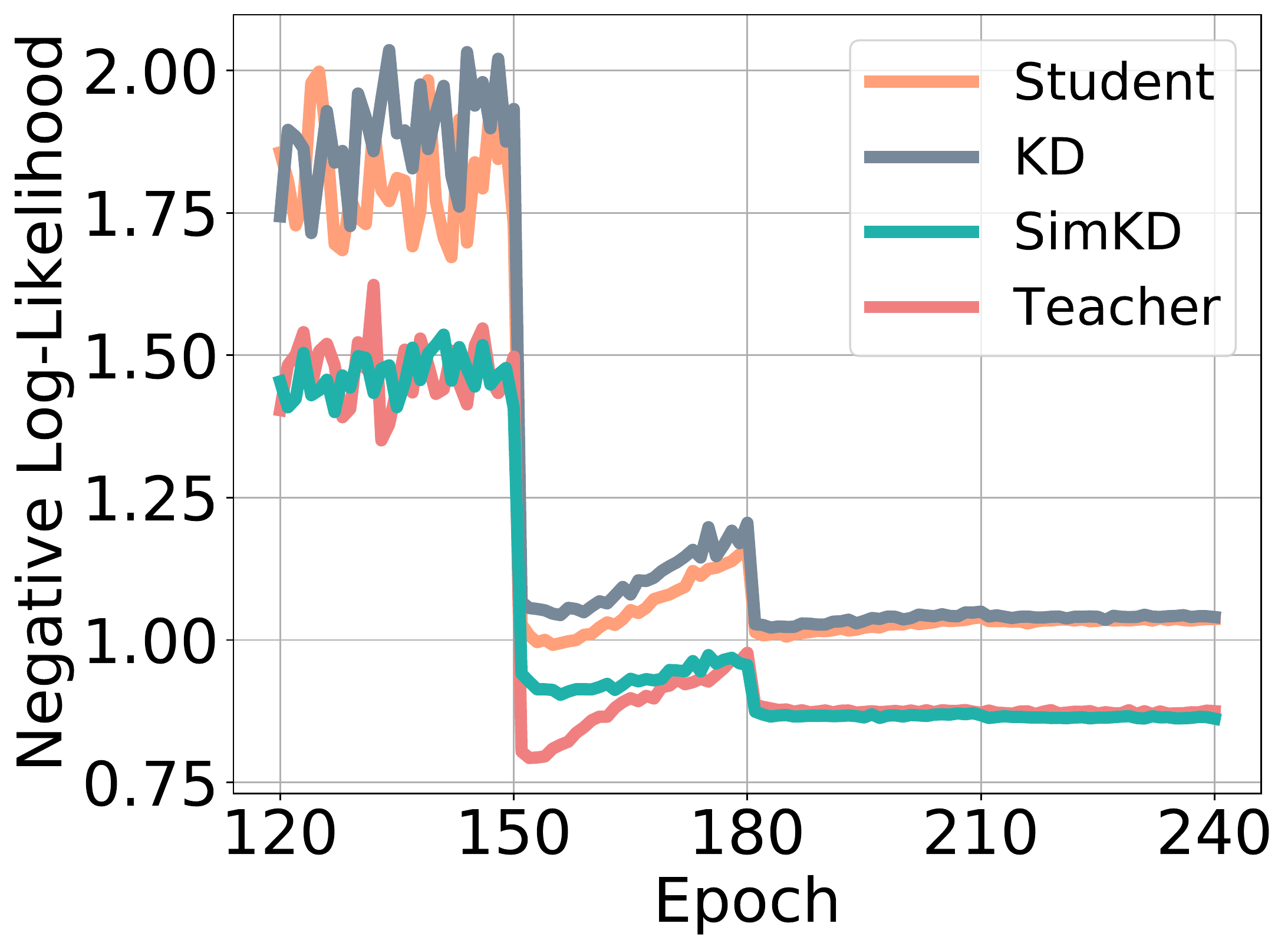}
	\end{minipage}
	\hfill
	\begin{minipage}{0.40\linewidth}
		\centering
		\resizebox{0.98\textwidth}{!}{
			\begin{tabular}{l|c}
				\toprule
				& Accuracy (\%) \\
				\midrule
				Student  		 & 73.09 $\pm$ 0.30 \\
				KD \cite{hinton2015distilling} & 74.42 $\pm$ 0.05 \\
				SimKD 			& \textbf{78.08 $\pm$ 0.15} \\
				SimKD+	      & \textbf{78.47 $\pm$ 0.08} \\
				SimKD++	    & \textbf{78.88 $\pm$ 0.05} \\
				\midrule
				Teacher 		& 79.42 \\
				\bottomrule
		\end{tabular}}
	\end{minipage}
	\caption{Comparison of the top-1 test accuracy (\%) and negative log-likelihood (Student: ResNet-8x4, Teacher: ResNet-32x4).}
	\label{fig:lossacc}
\end{figure}
\textbf{Reusing more teacher layers.}
We now generalize our technique to the situation where more deep layers of the teacher model are reused for student inference and show that the student performance will be further improved. 

We take ResNet architecture as an example and conduct experiments on CIAFR-100 dataset. Following the standard design, ResNet architecture consists of one convolutional layer, three building blocks and one fully-connected layer in a bottom-top fashion \cite{he2016deep}. Every building blocks contain the same number of convolutional layers and changing these layer numbers leads to different ResNet architectures. For example, 10 layers for each building block make up a 32-layer ResNet model. Then, besides reusing the final classifier as our SimKD do, two new variants are introduced by reusing additional last one or two building blocks, and they are denoted as ``SimKD+'' and ``SimKD++'', respectively. 

From Figure~\ref{fig:lossacc}, we can see that SimKD
significantly decreases negative log-likelihood by reusing only the teacher classifier, and its two variants further achieve higher performance as expected, though the associated complexity is also increased. These results support our hypothesis that reusing deep teacher layers is beneficial for the student performance improvement, probably due to most capability-specific information is contained in them.
Another explanation is that reusing more deep teacher layers would make the approximation of shallow teacher layers easier achievable and thus incur less performance degradation. In practice, reusing only the final teacher classifier strikes a good balance between performance and parameter complexity.

\subsection{Projector Analysis}

The parameter-free ``\textit{classifier-reusing}'' operation in our SimKD has been fully evaluated above. Next, we start to dig into another component\textemdash \textit{projector} from several aspects. We first present its default implementation and then show that it only requires a small number of extra parameters for achieving state-of-the-art performance. Finally, several ablation studies on the projector are provided.

\textbf{Implementation.} The aim of the projector $\mathcal{P}(\cdot)$ in Equation~(\ref{eq:simKD}) is to perfectly match the feature vectors $\boldsymbol{f}^{t}\in \mathbb{R}^{C_{t}}$ and $\boldsymbol{f}^s\in \mathbb{R}^{C_{s}}$. A na\"ive implementation is using one convolutional layer with batch normalization and ReLU activation, which has $C_{s}\times C_{t}+2\times C_{t}$ parameters \cite{yang2021knowledge}. However, this one-layer transformation may not suffice for accurate alignment due to the large capability gap between teacher and student models. We thus employ the last feature maps and a three-layer bottleneck transformation with dimension reduction factor $r$ as alternatives, hoping that these will help the features aligned better. The total parameters are 
\begin{equation}
	\label{eq:para}
	\begin{aligned}
		\frac{C_{t}(C_{s}+C_{t}+4)}{r}+\frac{9C_{t}^2}{r^2}+2C_{t}.
	\end{aligned}
\end{equation}
This formula implies that the added parameters will be reduced to between a quarter and a half if $r$ is doubled, which enables us to control the parameter complexity within an acceptable level by changing $r$. \textit{Detailed structure of the projector and analysis are provided in the technical appendix.} 

\begin{figure}[t]
	\centering
	\includegraphics[width=0.98\columnwidth]{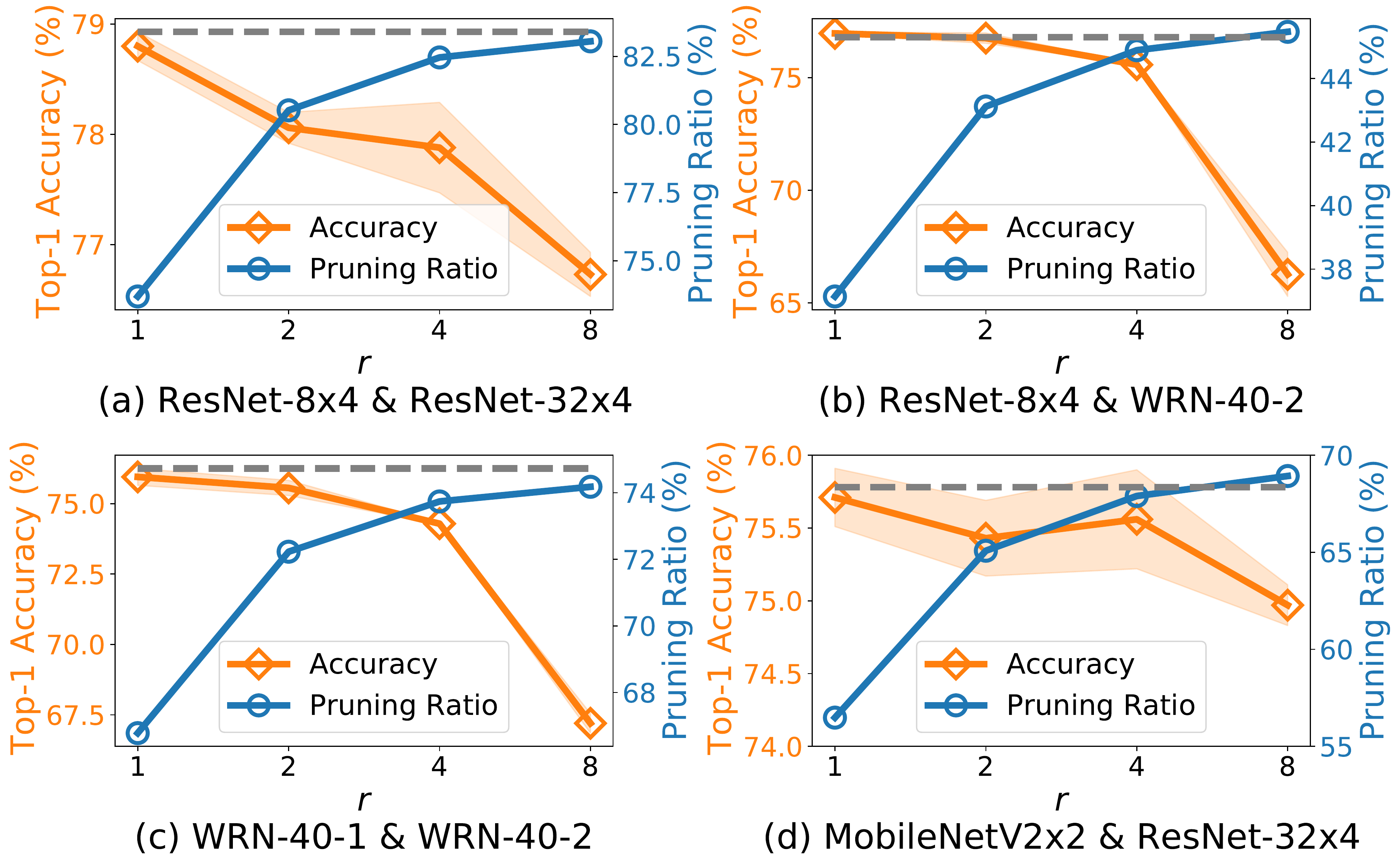}
	\caption{Trade-off between test accuracy and pruning ratio. The pruning ratio of the vanilla KD is drawn with the gray dashed line.}
	\label{fig:trade-off}
\end{figure}

\textbf{Effect to pruning ratio.} 
Figure~\ref{fig:trade-off} illustrates the trade-off between top-1 test accuracy and pruning ratio with different dimension reduction factor $r$. We adopt the following equation for the calculation of pruning ratio:
\begin{equation}
	\begin{aligned}
		\mbox{Pruning Ratio}&=1-\frac{\sharp \mbox{param}_{\mathrm{se}}+\sharp \mbox{param}_{\mathrm{proj}}+\Delta}{\sharp \mbox{param}_{\mathrm{t}}}\\
		\Delta&=\sharp\mbox{param}_{\mathrm{tc}}-\sharp\mbox{param}_{\mathrm{sc}}, 
	\end{aligned}
\end{equation}
where $\sharp \mbox{param}_{\mathrm{se}}$, $\sharp \mbox{param}_{\mathrm{proj}}$, $\sharp \mbox{param}_{\mathrm{t}}$ and $\sharp \mbox{param}_{\mathrm{tc/sc}}$ refer to the parameter number of a student encoder, a projector, a whole teacher model and a teacher/student classifier, respectively. Its upper bound is approached when $\sharp \mbox{param}_{\mathrm{proj}}\to 0$, which could be higher than the pruning ratio of the vanilla KD since $\sharp \mbox{param}_{\mathrm{proj}}+\Delta$ may be less than zero. Figure~\ref{fig:trade-off} shows that increasing $r$ will raise the pruning ratio, but in turn, cause the performance drop. This reduction may be attributed to that shrinking the bottleneck dimension of the projector will restrict its representation ability and thus affect the success of feature alignment. 
\begin{figure}
	\centering
	\includegraphics[width=0.85\columnwidth]{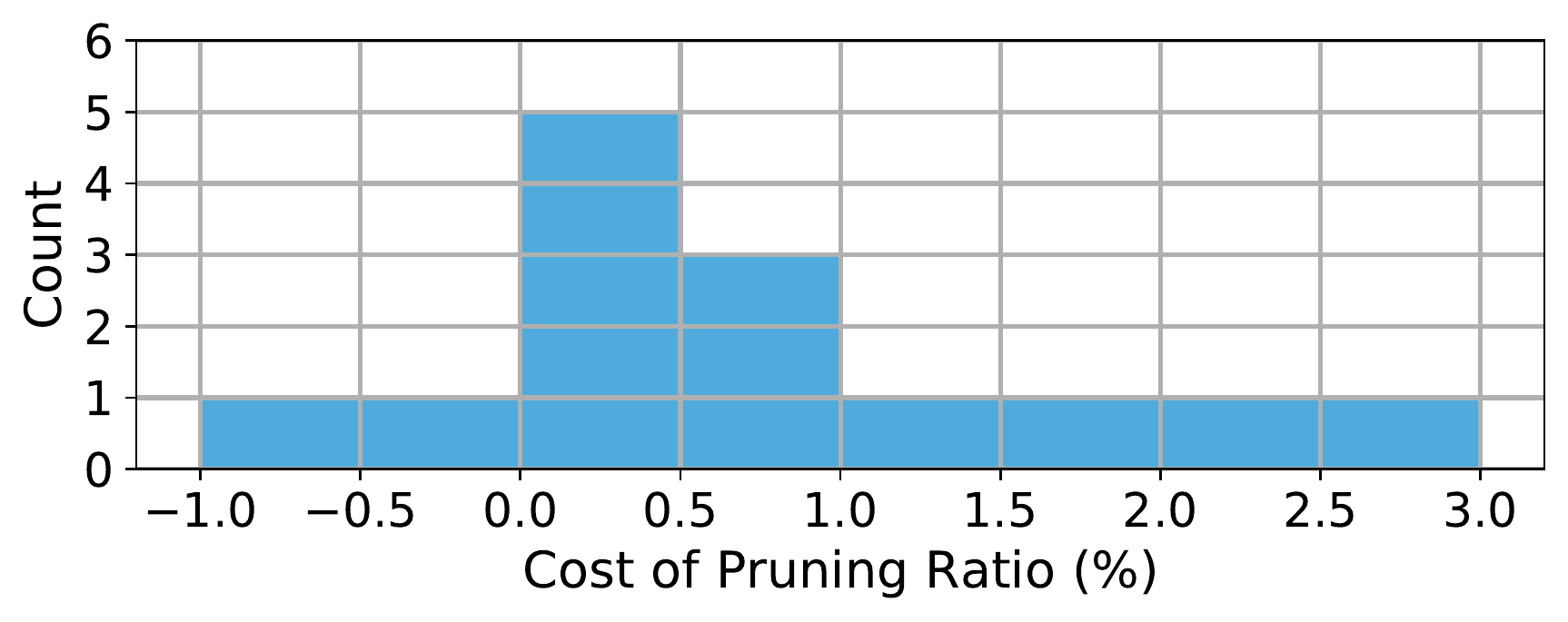}
	\caption{A histogram of the pruning ratio cost.}
	\label{fig:count}
\end{figure}

We then calculate the minimum pruning ratio cost of SimKD when it performs best in the competition on \textit{fourteen} teacher-student combinations from Table~\ref{Table:CIFAR-100} and~\ref{Table:CIFAR-100-2}. Figure~\ref{fig:count} show that our added projector only incurs less than 1\% pruning ratio cost in most cases (10/14). In some cases such as ``MobileNetV2x2 \& ResNet-32x4'' and ``ShuffleNetV1 \& ResNet-32x4'' with $r=8$, we find the pruning ratios of SimKD are even higher than the vanilla KD, and all competitors accordingly. Throughout this paper, we set $r$ equals 2 as default since this value strikes a good balance, i.e., gaining state-of-the-art results at the modest cost of pruning ratio. \textit{The full results are presented in the appendix.}

\textbf{Ablation study.}
We finally compare several implementations of the projector and loss function (see Appendix) for feature alignment. All results are obtained with the ``ResNet-8x4 \& ResNet-32x4'' combination on CIFAR-100. 

From Table~\ref{Table:proj}, the default implementation of our projector (the last row) achieves the best performance. The accuracy drop resulted from its simplified counterparts indicates the benefit of employing a relatively powerful projector in feature alignment. Moreover, the lower accuracy (76.03 $\pm$ 0.40) obtained by aligning feature vectors $\boldsymbol{f}^s$ and $\boldsymbol{f}^t$ with a three-layer transformation validates the effectiveness of using the last feature maps instead. Since our $\ell_2$ loss reflects the distance between extracted features, the lower test loss implies the closer alignment and thus the better test accuracy. This is consistent with the results in Table~\ref{Table:proj}. 

\begin{table}
	\centering
	\resizebox{0.96\columnwidth}{!}{
		\begin{tabular}{ccc}
			\toprule
			Projector  & Test loss ($\ell_2$) & Accuracy (\%) \\
			\midrule
			1x1Conv & 0.345 $\pm$ 0.001 & 75.15 $\pm$ 0.27 \\
			1x1Conv-1x1Conv & 0.343 $\pm$ 0.001 & 75.71 $\pm$ 0.33 \\
			1x1Conv-3x3Conv (DW)-1x1Conv & 0.306 $\pm$ 0.001 & 77.76 $\pm$ 0.12 \\
			1x1Conv-3x3Conv-1x1Conv & \textbf{0.301 $\pm$ 0.001} & \textbf{78.08 $\pm$ 0.15} \\
			\bottomrule
	\end{tabular}}
	\caption{Comparison of projectors. ``1x1/3x3Conv'' denotes a convolutional layer with 1x1/3x3 kernel size. ``DW'' denotes depthwise separable convolutions. Standard batch normalization and ReLU activation are used after each layer.}
	\label{Table:proj}
\end{table}

\subsection{Application \uppercase\expandafter{\romannumeral1}: Multi-Teacher Knowledge Distillation}

We then demonstrate the applicability of our technique in the multi-teacher KD setting where multiple pre-trained teacher models are available for the student training. 
Two representative approaches are compared: ``AVEG'' denotes a simple variant of the vanilla KD, which averages the predictions of multiple teachers; ``AEKD'' aggregates the teacher predictions with an adaptive weighting strategy and its improved version by incorporating intermediate features is denoted as ``AEKD-F'' \cite{du2020agree}.
As shown in Table~\ref{Table:multi-teacher}, SimKD always achieves the best performance. Additionally, we provide the results of $\text{SimKD}_{v}$, where a fully-connected layer projector is first used to align the feature vectors and then merged into the associated teacher classifier. The weights of multiple teacher classifiers are averaged and reused for student inference, which incurs no extra parameters.

\begin{table}
	\centering
	\resizebox{0.70\columnwidth}{!}{
		\begin{tabular}{ccc}
			\toprule
			Method  & {\large \ding{172}} & {\large \ding{173}} \\
			\midrule
			Student & 72.60 $\pm$ 0.12 & 72.60 $\pm$ 0.12 \\
			AVEG & 75.94 $\pm$ 0.20 & 76.33 $\pm$ 0.14  \\
			AEKD \cite{du2020agree} & 75.99 $\pm$ 0.18 & 76.17 $\pm$ 0.43 \\
			AEKD-F \cite{du2020agree} & 77.24 $\pm$ 0.32 & 77.08 $\pm$ 0.28 \\
			\midrule
			$\text{SimKD}_{v}$ & \textbf{77.43 $\pm$ 0.21} & \textbf{77.60 $\pm$ 0.23} \\
			SimKD & \textbf{78.59 $\pm$ 0.31} & \textbf{78.59 $\pm$ 0.05} \\
			\bottomrule
	\end{tabular}
}
	\caption{Results of the multi-teacher KD. We adopt ShuffleNetV2 as the student model and train it under two groups of pre-trained teacher models: {\large \ding{172}} includes three ResNet-32x4. {\large \ding{173}} includes two ResNet-32x4 and one ResNet-110x2.}
	\label{Table:multi-teacher}
\end{table}

\begin{table}
	\centering
	\resizebox{0.95\columnwidth}{!}{
	\begin{tabular}{cccc}
		\toprule
		Method  & Require data? & WRN-40-1 & WRN-16-2 \\
		\midrule
		Student & Yes & 71.92 $\pm$ 0.17 & 73.51 $\pm$ 0.32 \\
		\midrule
		ZSKT \cite{micaelli2019zskt} & No & 33.60 $\pm$ 3.88 & 45.03 $\pm$ 1.73  \\		
		DAFL \cite{chen2019dafl} & No & 45.32 $\pm$ 1.46 & 45.94 $\pm$ 1.66  \\
		CMI \cite{fang2021cmi} & No & 64.80 $\pm$ 0.35 & 65.11 $\pm$ 0.43 \\
		\midrule
		CMI+SimKD & No & \textbf{66.78 $\pm$ 0.29} & \textbf{67.31 $\pm$ 0.89} \\
		\bottomrule
	\end{tabular}
	}
	\caption{Results of the data-free KD. We adopt WRN-40-2 as the teacher model with two different student models.}
	\label{Table:data-free}
\end{table}

\subsection{Application \uppercase\expandafter{\romannumeral2}: Data-Free Knowledge Distillation}
Data-free knowledge distillation aims to exploit a pre-trained teacher model without accessing its training dataset to improve the student performance. A popular paradigm is to recover the original data manifold with a generative model first and then perform knowledge distillation on the synthesized dataset \cite{micaelli2019zskt,chen2019dafl,fang2021cmi}. Our SimKD can be easily integrated into these existing approaches by replacing their KD training step as our ``reusing-classifier'' operation and the associated feature alignment. Table~\ref{Table:data-free} shows that with the help of our SimKD, the student performance is also improved in the data-free knowledge distillation application.

\section{Conclusion}
In this paper, we have explored a simple knowledge distillation technique where the pre-trained teacher classifier is reused for student inference and the student model is trained with a single $\ell_2$ loss for feature alignment.
We design several experiments to analyze the workings of our technique and conduct extensive experiments to demonstrate its superiority over state-of-the-art approaches. We hope this study will be an important baseline for future research.

\section{Limitation and Future Work}
A simple parameter reusing is served as our first attempt to explore the potential value of the teacher classifier. This requires a projector when feature dimensions are mismatched and thus increases the model complexity. 
How to develop a projector-free alternative needs further investigation.
Another limitation is that our technique is only applicable for supervised knowledge distillation, such as image classification \cite{hinton2015distilling}, dense prediction \cite{shu2021cwkd} and machine translation \cite{tan2019multilingualNMT}. It is also worthwhile to develop a successful variant of our technique for unsupervised learning scenario. 

\section{Acknowledgment}
This work is supported by the Starry Night Science Fund of Zhejiang University Shanghai Institute for Advanced Study (Grant No: SN-ZJU-SIAS-001), National Natural Science Foundation of China (Grant No: U1866602) and Zhejiang Provincial Natural Science Foundation (Grant No: LY20F020023). The authors would like to thank Chunhua Shen and anonymous reviewers for their helpful comments.

\appendix
\renewcommand\thefigure{\thesection.\arabic{figure}}
\renewcommand\thetable{\thesection.\arabic{table}}
\setcounter{figure}{0} 
\setcounter{table}{0} 

\setcounter{figure}{0}
\renewcommand{\thefigure}{S.\arabic{figure}}

\setcounter{table}{0}
\renewcommand{\thetable}{S.\arabic{table}}
\section{Experimental Setting}
\subsection{Datasets and Training Details}
We adopt two datasets including CIFAR-100 \cite{krizhevsky2009learning} and ImageNet \cite{russakovsky2015ImageNet} to conduct experiments. All images are normalized by channel means and standard deviations. A horizontal flip is used for data augmentation. 
\textbf{CIFAR-100\footnote{\url{https://www.cs.toronto.edu/~kriz/cifar.html}}} contains 50,000 training images and 10,000 test images from 100 classes. Each training image is padded by 4 pixels on each size and randomly cropped as a $32\times32$ sample. 
\textbf{ImageNet\footnote{\url{http://image-net.org/challenges/LSVRC/2012/index}}} contains about 1.3 million training images and 50,000 validation images from 1,000 classes. Each image is randomly cropped as a 224x224 sample without padding. The top-1 test accuracy of the teacher model (ResNet-50) is 76.26\%.

\textbf{Multi-Teacher Knowledge Distillation.}
The training hyper-parameters of multi-teacher KD are exactly the same as those of single-teacher KD on CIFAR-100. We first pre-train multiple teacher models with different initialization and then distill their knowledge into a student model. The accuracies of compared AEKD and AEKD-F \cite{du2020agree} are obtained by running a public library\footnote{\url{https://github.com/Rorozhl/CA-MKD}} with default model hyper-parameters on our teacher-student combinations \cite{zhang2021confidence}. The top-1 test accuracy of two groups of teacher models used in our main submission are: {\large \ding{172}} Three ResNet-32x4 models (79.32, 79.43, 79.45), {\large \ding{173}} Two ResNet-32x4 models (79.43, 79.45) and one ResNet-110x2 model (78.18).

\textbf{Data-Free Knowledge Distillation.} 
We adopt a public library\footnote{\url{https://github.com/zju-vipa/DataFree}} to reproduce the results of compared approaches: ZSKT \cite{micaelli2019zskt}, DAFL \cite{chen2019dafl} and CMI \cite{fang2021cmi}, with the default model hyper-parameters. In our experiment, the top-1 test accuracy of the teacher model (WRN-40-2) is 76.31\%. The performance of the student model trained with original dataset is included for comparison.

\textbf{Computing Infrastructure.} All of the experiments are conducted with PyTorch \cite{paszke2019pytorch}. CIFAR-100 experiments are conducted on a sever containing eight NVIDIA GeForce RTX 2080Ti GPUs with 11GB RAM. The CUDA version is 11.2. 
ImageNet experiments are conducted on a sever containing four NVIDIA A40 GPUs with 48GB RAM. The CUDA version is 11.4. 

\subsection{Network Architectures}
We use a large number of teacher-student combinations for performance evaluation, which are composed of several popular neural network architectures: VGG \cite{simonyan2014very}, ResNet \cite{he2016deep}, WRN \cite{zagoruyko2016wide}, MobileNetV2 \cite{sandler2018mobile}, ShuffleNetV1 \cite{zhang2018shufflenet}, ShuffleNetV2 \cite{ma2018shuffle}. The number behind ``VGG-'', ``ResNet-'' denotes the depth of networks. ``WRN-d-w'' denotes the wide-ResNet with depth $d$ and width factor $w$. 
As the previous works do \cite{tian2020contrastive,chen2021cross}, we expand or shrink the number of convolution filters in intermediate layers of some network architectures with a certain ratio and put that ratio behind ``x'', such as ``ResNet-32x4''. 
\begin{table}
	\centering
	\begin{tabular}{ccc}
		\toprule
		Input dimension & Operator & Output dimension \\
		\midrule
		$H\times W\times C_{s}$ & 1x1 Conv  & $H\times W\times C_{t}/r$ \\
		$H\times W\times C_{t}/r$ & 3x3 Conv  & $H\times W\times C_{t}/r$ \\
		$H\times W\times C_{t}/r$ & 1x1 Conv & $H\times W\times C_{t}$ \\
		\bottomrule
	\end{tabular}
	\caption{Projector structure. ``1x1/3x3Conv'' denotes a convolutional layer with 1x1/3x3 kernel size. Standard batch normalization and ReLU activation are used after each convolutional layer.  $r$ is the reduction ratio.}
	\label{Table:app_proj}
\end{table}

\subsection{Projector}

The detailed structure of our used projector is described in Table~\ref{Table:app_proj}. We assume that the spatial dimensions of involved feature maps are the same, and denote them with the notations $H$ and $W$. Otherwise, an average pooling operation is used in advance for spatial dimension alignment to reduce the computational consumption, as the previous work do \cite{chen2021cross}.

Given the feature maps of teacher and student models, the parameter number of the added projector is a function of the dimension reduction factor $r$

\begin{equation}
	\label{eq:ap-para}
	\begin{aligned}
		\mathcal{F}(r)=\frac{C_{t}(C_{s}+C_{t}+4)}{r}+\frac{9C_{t}^2}{r^2}+2C_{t}.\\
	\end{aligned}
\end{equation}

\textbf{Proposition.} The extra parameter number $\mathcal{F}(r)$ satisfies the inequality  $2\mathcal{F}(2r)<\mathcal{F}(r)< 4\mathcal{F}(2r)$ under some mild conditions.

\textbf{Proof.} \\
We first prove the left part of the inequality:
\begin{equation}
	\begin{aligned}
		2\mathcal{F}(2r)&<\mathcal{F}(r)\\			2\times\left(\frac{C_{t}(C_{s}+C_{t}+4)}{2r}+\frac{9C_{t}^2}{4r^2}+2C_{t}\right)
		&<\\
		\frac{C_{t}(C_{s}+C_{t}+4)}{r}+\frac{9C_{t}^2}{r^2}+2C_{t}\\
		2\times\left(\frac{9C_{t}^2}{4r^2}+2C_{t}\right)
		&<\frac{9C_{t}^2}{r^2}+2C_{t}\\
		2C_{t}-\frac{9C_{t}^2}{2r^2}
		&<0\\
		2C_{t}\left(1-\frac{9C_{t}}{4r^2}\right)
		&<0.
	\end{aligned}
\end{equation}
Generally, the channel dimension $C_t$ in the last feature maps of popular deep neural networks is greater than 128 on CIFAR-100 and is greater than 512 on ImageNet, which means that this equation holds when $r<16$ and $r<32$, respectively. This is easy to be satisfied in practice. Since 
a typical setting for $r$ is 1, 2 and 4 in order to avoid substantial accuracy reduction as shown in Table~\ref{Table:app_pruning-1} and~\ref{Table:app_pruning-2}. 

We then prove the right part of the inequality:
\begin{equation}
	\begin{aligned}
		4\mathcal{F}(2r)&>\mathcal{F}(r)\\
		4\times\left(\frac{C_{t}(C_{s}+C_{t}+4)}{2r}+\frac{9C_{t}^2}{4r^2}+2C_{t}\right)
		&>\\\frac{C_{t}(C_{s}+C_{t}+4)}{r}+\frac{9C_{t}^2}{r^2}+2C_{t}\\
		4\times\left(\frac{C_{t}(C_{s}+C_{t}+4)}{2r}+2C_{t}\right)
		&>\\\frac{C_{t}(C_{s}+C_{t}+4)}{r}+2C_{t}\\
		2\times\left(\frac{C_{t}(C_{s}+C_{t}+4)}{2r}+3C_{t}\right)
		&>0.
	\end{aligned}
\end{equation}
Since the channel dimensions $C_t$ and $C_s$ are always greater than zero, this inequality holds automatically.
\qed

\begin{table}[t]
	\centering
	\resizebox{0.9\columnwidth}{!}{
		\begin{tabular}{c|ccc}
			\toprule
			\multirow{2}{*}{Student} & VGG-8 & WRN-16-2 & WRN-16-4 \\
			& 70.46 $\pm$ 0.29 & 73.51 $\pm$ 0.32 & 77.26 $\pm$ 0.24 \\
			\midrule
			KD \cite{hinton2015distilling}  & 73.38 $\pm$ 0.05 & 75.40 $\pm$ 0.34 & 79.24 $\pm$ 0.23\\
			FitNet \cite{romero2015fitnets} & 73.63 $\pm$ 0.11 & 75.44 $\pm$ 0.22 & 79.06 $\pm$ 0.16 \\
			AT \cite{zagoruyko2017paying} & 73.51 $\pm$ 0.08 & 75.76 $\pm$ 0.29 & 79.38 $\pm$ 0.20 \\
			SP \cite{tung2019similarity} & 73.53 $\pm$ 0.23 & 75.61 $\pm$ 0.34 & 79.53 $\pm$ 0.20  \\			
			VID \cite{ahn2019variational}  & 73.63 $\pm$ 0.07 & 75.44 $\pm$ 0.24 & 79.40 $\pm$ 0.08\\
			CRD \cite{tian2020contrastive}  & 74.31 $\pm$ 0.17 & 75.86 $\pm$ 0.17 & 79.46 $\pm$ 0.19 \\
			SRRL \cite{yang2021knowledge} & 74.25 $\pm$ 0.35 & 75.89 $\pm$ 0.12 & 79.67 $\pm$ 0.17  \\
			SemCKD \cite{chen2021cross}  & 74.43 $\pm$ 0.25 & 75.77 $\pm$ 0.11 & 80.05 $\pm$ 0.27\\
			\midrule
			SimKD & \textbf{74.93 $\pm$ 0.21} & \textbf{76.23 $\pm$ 0.14} & \textbf{80.36 $\pm$ 0.04} \\
			\midrule
			\multirow{2}{*}{Teacher}  &VGG-13 & WRN-40-2& WRN-40-4 \\
			& 74.64 & 76.31 & 79.51 \\
			\bottomrule
		\end{tabular}
	}
	\caption{Top-1 test accuracy (\%) comparison on CIFAR-100.}
	\label{Table:app_cifar}
\end{table}  

\begin{table}
	\centering
	\resizebox{0.96\columnwidth}{!}{
		\begin{tabular}{cccc}
			\toprule
			Network & Student & SimKD & Teacher \\
			\midrule
			ResNet-34 \& ResNet-50 & 74.01 & \textbf{74.64} & 76.26 \\
			ResNet-50 \& ResNet-101 & 76.26 & \textbf{77.60} & 77.80 \\
			\bottomrule
		\end{tabular}
	}
	\caption{Top-1 test accuracy (\%) comparison on ImageNet. }
	\label{Table:app_imagenet}
\end{table}   
\begin{figure}
	\centering
	\includegraphics[width=0.82\columnwidth]{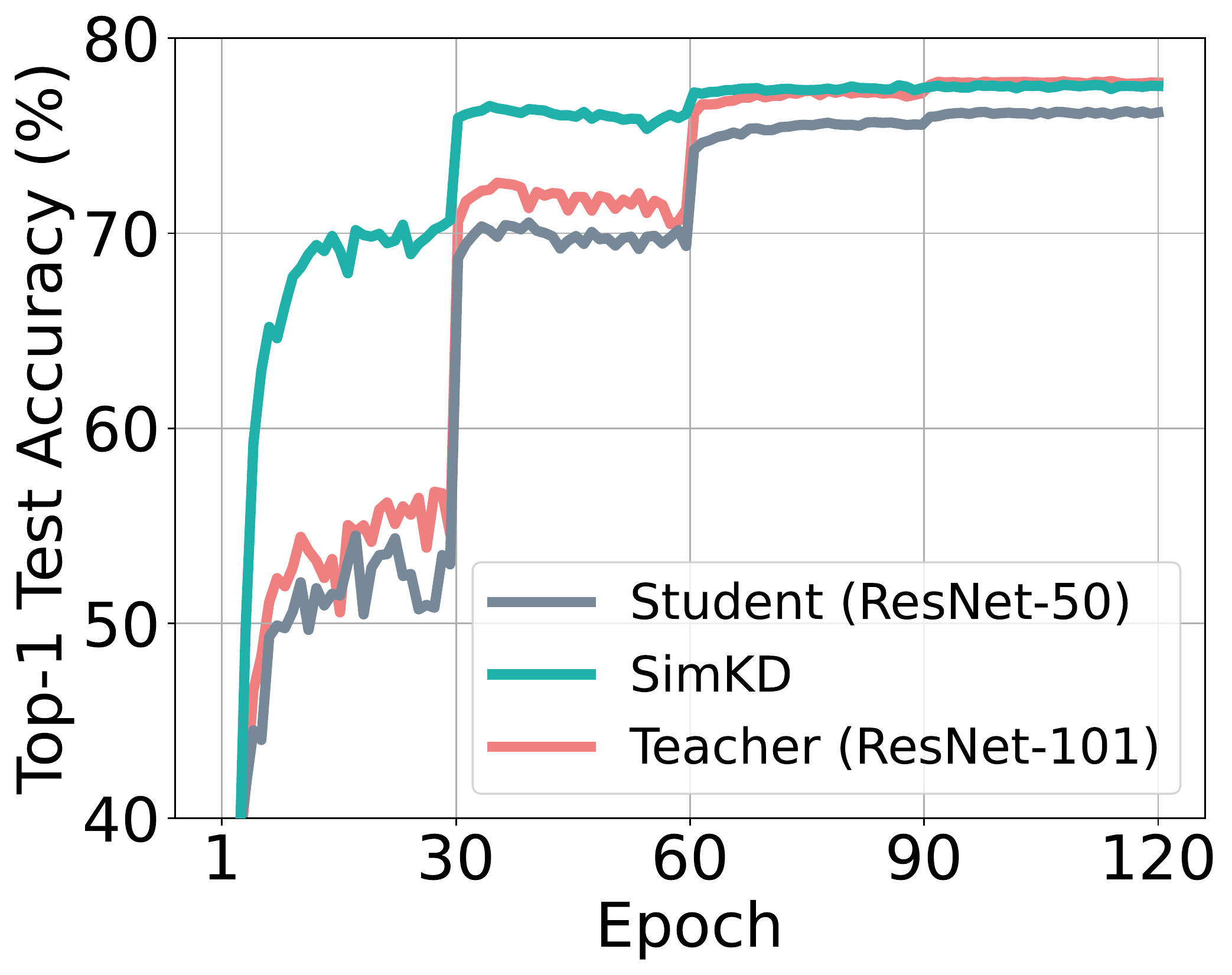}
	\caption{The test accuracy (\%) of ResNet-50 \& ResNet-101 on ImageNet. Our SimKD achieves faster model convergence.}
	\label{fig:imagenet_acc}
\end{figure}

\begin{table*}[t]
	\centering
	\begin{tabular}{l|cc|cc}
		\toprule
		\multirow{2}{*}{Student} & \multicolumn{2}{c|}{ResNet-8x4} & \multicolumn{2}{c}{ResNet-8x4} \\
		& \multicolumn{2}{c|}{73.09 $\pm$ 0.30} & \multicolumn{2}{c}{73.09 $\pm$ 0.30} \\
		\midrule
		& Student Classifier & Teacher Classifier & Student Classifier & Teacher Classifier \\
		\midrule
		$\alpha=0$ (KD) & \textbf{74.42 $\pm$ 0.05} & -- & \textbf{75.28 $\pm$ 0.18} & -- \\
		\midrule
		$\alpha=0.2$ & 74.42 $\pm$ 0.10 & 73.91 $\pm$ 0.12 & 74.83 $\pm$ 0.29 & 73.45 $\pm$ 0.31 \\
		$\alpha=0.4$ & 73.99 $\pm$ 0.03 & 73.83 $\pm$ 0.13 & 74.72 $\pm$ 0.17 & 73.65 $\pm$ 0.27 \\
		$\alpha=0.6$ & 73.93 $\pm$ 0.08 & 73.89 $\pm$ 0.26 & 74.59 $\pm$ 0.23 & 73.81 $\pm$ 0.25 \\
		$\alpha=0.8$ & 73.76 $\pm$ 0.26 & 74.13 $\pm$ 0.24 & 74.29 $\pm$ 0.24 & 74.19 $\pm$ 0.22 \\
		$\alpha=0.9$ & 73.52 $\pm$ 0.33 & 74.42 $\pm$ 0.26 & 73.98 $\pm$ 0.06 & 74.73 $\pm$ 0.13 \\
		$\alpha=0.99$ & 30.77 $\pm$ 0.92 & 77.66 $\pm$ 0.22 & 23.46 $\pm$ 0.81 & 76.68 $\pm$ 0.13 \\
		$\alpha=0.999$ & 5.59 $\pm$ 0.60 & \textbf{77.98 $\pm$ 0.19} & 4.55 $\pm$ 0.68 & \textbf{76.84 $\pm$ 0.28} \\
		\midrule
		$\alpha=1$ (SimKD) & -- & \textbf{78.08 $\pm$ 0.15}& -- & \textbf{76.75 $\pm$ 0.23} \\
		\midrule
		\multirow{2}{*}{Teacher} & \multicolumn{2}{c|}{ResNet-32x4} & \multicolumn{2}{c}{WRN-40-2} \\
		& \multicolumn{2}{c|}{79.42} & \multicolumn{2}{c}{76.31} \\
		\bottomrule
	\end{tabular}
	\caption{Joint training the student feature encoder and classifier with different hyper-parameters.}
	\label{Table:app_joint-1}
\end{table*}  

\begin{table*}[t]
	\centering
	\begin{tabular}{l|cc|cc}
		\toprule
		\multirow{2}{*}{Student} & \multicolumn{2}{c|}{WRN-40-1} & \multicolumn{2}{c}{MobileNetV2x2} \\
		& \multicolumn{2}{c|}{71.92 $\pm$ 0.17} & \multicolumn{2}{c}{69.06 $\pm$ 0.10} \\
		\midrule
		& Student Classifier & Teacher Classifier & Student Classifier & Teacher Classifier \\
		\midrule
		$\alpha=0$ (KD) & \textbf{74.12 $\pm$ 0.29} & -- & \textbf{72.43 $\pm$ 0.32} & -- \\
		\midrule
		$\alpha=0.2$ & 74.33 $\pm$ 0.32 & 73.92 $\pm$ 0.31 & 72.50 $\pm$ 0.26 & 72.04 $\pm$ 0.23 \\
		$\alpha=0.4$ & 74.49 $\pm$ 0.30 & 74.17 $\pm$ 0.19 & 72.48 $\pm$ 0.38 & 72.13 $\pm$ 0.25 \\
		$\alpha=0.6$ & 74.53 $\pm$ 0.11 & 74.39 $\pm$ 0.20 & 72.83 $\pm$ 0.34 & 72.64 $\pm$ 0.29 \\
		$\alpha=0.8$ & 74.93 $\pm$ 0.25 & 74.88 $\pm$ 0.22 & 73.13 $\pm$ 0.15 & 72.95 $\pm$ 0.06 \\
		$\alpha=0.9$ & 75.18 $\pm$ 0.20 & 75.17 $\pm$ 0.17 & 73.44 $\pm$ 0.26 & 73.40 $\pm$ 0.27 \\
		$\alpha=0.99$ & 74.17 $\pm$ 0.06 & 75.35 $\pm$ 0.15 & 74.91 $\pm$ 0.30 & 75.31 $\pm$ 0.16 \\
		$\alpha=0.999$ & 18.39 $\pm$ 1.45 & 75.33 $\pm$ 0.08 & 13.49 $\pm$ 1.80 & \textbf{75.43 $\pm$ 0.22} \\
		\midrule
		$\alpha=1$ (SimKD) & -- & \textbf{75.56 $\pm$ 0.27} & -- & \textbf{75.43 $\pm$ 0.26} \\
		\midrule
		\multirow{2}{*}{Teacher} & \multicolumn{2}{c|}{WRN-40-2} & \multicolumn{2}{c}{ResNet-32x4} \\
		& \multicolumn{2}{c|}{76.31} & \multicolumn{2}{c}{79.42} \\
		\bottomrule
	\end{tabular}
	\caption{Joint training the student feature encoder and classifier with different hyper-parameters.}
	\label{Table:app_joint-2}
\end{table*}
\section{More Experimental Results}
\subsection{Comparison of Test Accuracy}
Table~\ref{Table:app_cifar} and~\ref{Table:app_imagenet} presents more results on CIFAR-100 and ImageNet datasets with extra \textit{five} teacher-student combinations. Similar observations are obtained as those in the main submission. For ImageNet dataset, we replace the 3x3 convolution as the 3x3 depth-wise separable convolution in the projector (Table~\ref{Table:app_proj}) to control the extra parameters. 

As shown in Figure~\ref{fig:imagenet_acc}, our SimKD achieves \textit{faster convergence} in the whole model training. For example, at 30th epoch, SimKD performance is on the par with the baseline student model performance at 60th epoch. Besides, at 60th epoch, SimKD already outperforms the baseline student model at 120th epoch.

\subsection{Joint Training Results}
Table~\ref{Table:app_joint-1} and~\ref{Table:app_joint-2} present the full joint training results with different hyper-parameters. 
In the case of $\alpha=0$ or $\alpha=1$, only the student classifier or teacher classifier produces meaningful results and the another one degrades into random guess. We denote these random guess as ``--''. 

\begin{table}[t]
	\centering
	\begin{tabular}{l|c}
		\toprule
		Learning Rate & Test Accuracy \\
		\midrule
		0.01 & 52.03 $\pm$ 0.15 \\
		\textbf{0.05} & \textbf{51.97 $\pm$ 0.19}  \\
		0.1 & 52.01 $\pm$ 0.17 \\
		0.5 & 51.93 $\pm$ 0.20 \\
		\bottomrule
	\end{tabular}
	\caption{Training a new classifier from scratch with different initial learning rates (Student: ResNet-8x4, Teacher: ResNet-32x4).}
	\label{Table:app_seq}
\end{table}

\begin{table*}[t]
	\centering
	\begin{tabular}{cccc}
		\toprule
		& Input ($\ell_2$) & Output ($\ell_2$) & Input ($\ell_2$) + Output ($\ell_2$) \\
		\midrule
		Accuracy & \textbf{78.08 $\pm$ 0.15} & 77.09 $\pm$ 0.09 & 77.88 $\pm$ 0.30 \\
		\midrule
		Loss function &$ \|\boldsymbol{f}^t - \boldsymbol{f}^s\|^2_2$ & $\|\boldsymbol{W}^t\boldsymbol{f}^t - \boldsymbol{W}^t\boldsymbol{f}^s\|^2_2$ & $\|\boldsymbol{f}^t - \boldsymbol{f}^s\|^2_2 				+\|\boldsymbol{W}^t\boldsymbol{f}^t - \boldsymbol{W}^t\boldsymbol{f}^s\|^2_2$ \\
		\midrule
		Gradient on $\boldsymbol{f}^s$ & $-2\left(\boldsymbol{f}^t - \boldsymbol{f}^s\right)$ & $-2{\mathbf{W}^t}^\mathrm{T}\mathbf{W}^t\left(\boldsymbol{f}^t - \boldsymbol{f}^s\right)$& $-2\{ (\boldsymbol{I}+{\boldsymbol{W}^t}^\mathrm{T}\boldsymbol{W}^t)(\boldsymbol{f}^t - \boldsymbol{f}^s)\}$\\
		\bottomrule
	\end{tabular}
	\caption{Comparison of different loss functions (Student: ResNet-8x4, Teacher: ResNet-32x4). We omit the projector $\mathcal{P}(\cdot)$ for simplicity.}
	\label{Table:app_loss}
\end{table*}

\begin{table*}[htbp]
	\centering
	\resizebox{0.99\textwidth}{!}{
		\begin{tabular}{c|ccccccc}
			\toprule
			\multirow{2}{*}{Student} &  WRN-40-1 & ResNet-8x4  & ResNet-110 & ResNet-116 & VGG-8 & ResNet-8x4 & ShuffleNetV2 \\
			& 71.92 $\pm$ 0.17 & 73.09 $\pm$ 0.30 & 74.37 $\pm$ 0.17 & 74.46 $\pm$ 0.09 & 70.46 $\pm$ 0.29 & 73.09 $\pm$ 0.30 & 72.60 $\pm$ 0.12  \\
			\midrule
			\multirow{2}{*}{$r=8$} & 67.20 $\pm$ 0.35 & \textbf{76.73 $\pm$ 0.20} & 71.71 $\pm$ 1.00 & 71.96 $\pm$ 1.09 & \textbf{74.74 $\pm$ 0.15} & 66.26 $\pm$ 0.98 & 77.49 $\pm$ 0.31 \\
			& (0.55\%, 1.05\%) & (\textbf{0.35\%}, 2.11\%) & (0.18\%, 0.35\%) & (0.18\%, 0.33\%) & (\textbf{0.90\%}, 0.86\%) & (-0.17\%, 0.73\%) & (-0.35\%, 3.76\%) \\
			\midrule
			\multirow{2}{*}{$r=4$} & 74.29 $\pm$ 0.03 & \textbf{77.88 $\pm$ 0.41} & \textbf{77.14 $\pm$ 0.22} & \textbf{77.18 $\pm$ 0.21} & \textbf{75.62 $\pm$ 0.28} & 75.57 $\pm$ 0.03 & \textbf{78.21 $\pm$ 0.20} \\			
			& (0.99\%, 2.81\%) & (\textbf{0.94\%}, 5.67\%) & (\textbf{0.32\%}, 0.92\%) & (\textbf{0.32\%}, 0.87\%) & (\textbf{1.62\%}, 2.19\%) & (0.41\%, 1.78\%) & (\textbf{0.58\%}, 8.85\%) \\	
			\midrule		
			\multirow{2}{*}{$r=2$} & \textbf{75.56 $\pm$ 0.27} & \textbf{78.08 $\pm$ 0.15}  & \textbf{77.82 $\pm$ 0.15} & \textbf{77.90 $\pm$ 0.11} & \textbf{75.76 $\pm$ 0.12} & \textbf{76.75 $\pm$ 0.23} & \textbf{78.39 $\pm$ 0.27} \\
			& (\textbf{2.5\%}, 8.77\%) & (\textbf{2.88\%}, 17.34\%) & (\textbf{0.82\%}, 2.88\%) & (\textbf{0.82\%}, 2.73\%) & (\textbf{2.98\%}, 6.23\%) & (\textbf{2.18\%}, 5.02\%) & (\textbf{3.16\%}, 23.01\%) \\
			\midrule
			\multirow{2}{*}{$r=1$} & 75.95 $\pm$ 0.30 & 78.80 $\pm$ 0.13 & \textbf{78.00 $\pm$ 0.26} & \textbf{78.15 $\pm$ 0.30} & 75.98 $\pm$ 0.21 & 76.96 $\pm$ 0.07 & 78.66 $\pm$ 0.08 \\
			& (7.95\%, 30.35\%) & (9.71\%, 58.51\%) & (\textbf{2.6\%}, 9.96\%) & (\textbf{2.6\%}, 9.43\%) & (11.05\%, 19.87\%) & (8.16\%, 15.96\%) & (11.33\%, 67.77\%) \\
			\midrule
			\multirow{2}{*}{Teacher} & WRN-40-2 & ResNet-32x4 & ResNet-110x2 & ResNet-110x2 & ResNet-32x4 & WRN-40-2 & ResNet-32x4  \\
			& 76.31 & 79.42 & 78.18 & 78.18 & 79.42 & 76.31 & 79.42  \\
			\bottomrule
		\end{tabular}
	}
	\caption{Top-1 test accuracy (\%) and pruning ratio (the first element in parenthesis) of SimKD with various dimension reduction factor $r$ on CIFAR-100. We also provide the ratio of the projector parameters to the student parameters (the second element in parenthesis).}
	\label{Table:app_pruning-1}
\end{table*} 
\begin{table*}[t]
	\centering
	\resizebox{0.99\textwidth}{!}{
		\begin{tabular}{c|ccccccc}
			\toprule
			\multirow{2}{*}{Student} & ShuffleNetV1 & WRN-16-2 & ShuffleNetV2 & MobileNetV2 & MobileNetV2x2 & WRN-40-2  & ShuffleNetV2x1.5 \\
			& 71.36 $\pm$ 0.25 & 73.51 $\pm$ 0.32 & 72.60 $\pm$ 0.12 & 65.43 $\pm$ 0.29 & 69.06 $\pm$ 0.10 & 76.35 $\pm$ 0.18 & 74.15 $\pm$ 0.22 \\
			\midrule
			\multirow{2}{*}{$r=8$} & \textbf{76.68 $\pm$ 0.20} & 75.41 $\pm$ 0.17 & 73.50 $\pm$ 0.77 & 61.78 $\pm$ 1.21 & \textbf{74.97 $\pm$ 0.14} & 78.55 $\pm$ 0.33 & 78.96 $\pm$ 0.15 \\
			& (\textbf{-0.29\%}, 5.16\%) & (0.47\%, 3.13\%) & (-0.99\%, 1.55\%) & (-4.00\%, 3.08\%) & (\textbf{-0.58\%}, 2.51\%) & (0.13\%, 0.98\%) & (-0.35\%, 1.98\%) \\
			\midrule
			\multirow{2}{*}{$r=4$} & \textbf{77.22 $\pm$ 0.22} & \textbf{76.69 $\pm$ 0.23} & 77.35 $\pm$ 0.18 & 69.43 $\pm$ 0.21 & \textbf{75.56 $\pm$ 0.34} & \textbf{79.23 $\pm$ 0.06} & \textbf{79.48 $\pm$ 0.12} \\
			& (\textbf{0.60\%}, 12.12\%) & (\textbf{1.01\%}, 8.81\%) & (-0.63\%, 3.39\%) & (-2.67\%, 6.77\%) & (\textbf{0.45\%}, 5.78\%) & (\textbf{0.27\%}, 2.75\%) & (\textbf{0.58\%}, 4.65\%) \\
			\midrule
			\multirow{2}{*}{$r=2$} & \textbf{77.18 $\pm$ 0.26} & \textbf{77.17 $\pm$ 0.32} & \textbf{78.25 $\pm$ 0.24} & \textbf{70.71 $\pm$ 0.41} & \textbf{75.43 $\pm$ 0.26} & \textbf{79.29 $\pm$ 0.11} & \textbf{79.54 $\pm$ 0.26} \\
			& (\textbf{3.14\%}, 32.03\%) & (\textbf{2.84\%}, 28.13\%) & (\textbf{0.31\%}, 8.19\%) & (\textbf{0.52\%}, 15.62\%) & (\textbf{3.26\%}, 14.66\%) & (\textbf{0.76\%}, 8.78\%) & (\textbf{3.16\%}, 12.09\%) \\
			\midrule
			\multirow{2}{*}{$r=1$} & 77.58 $\pm$ 0.36 & 77.65 $\pm$ 0.24 & 78.58 $\pm$ 0.22 & 70.90 $\pm$ 0.17 & 75.71 $\pm$ 0.20 & 79.26 $\pm$ 0.17 & 79.72 $\pm$ 0.24 \\
			& (11.20\%, 95.15\%) & (9.45\%, 98.01\%) & (2.99\%, 21.83\%) & (9.43\%, 40.34\%) & (11.87\%, 41.86\%) & (2.55\%, 30.59\%) & (11.33\%, 35.61\%) \\
			\midrule
			\multirow{2}{*}{Teacher} & ResNet-32x4 & ResNet-32x4 & ResNet-110x2 & WRN-40-2 & ResNet-32x4 & ResNet-110x4 & ResNet-32x4 \\
			& 79.42 & 79.42 & 78.18 & 76.31 & 79.42 & 80.20 & 79.42 \\
			\bottomrule
		\end{tabular}
	}
	\caption{Top-1 test accuracy (\%) and pruning ratio (the first element in parenthesis) of SimKD with various dimension reduction factor $r$ on CIFAR-100. We also provide the ratio of the projector parameters to the student parameters (the second element in parenthesis).}
	\label{Table:app_pruning-2}
\end{table*}  

\subsection{Sequential Training Results}
The results of sequential training in the main submission are obtained with the regular training procedure. That is to say, we adopt SGD with 0.9 Nesterov momentum and $5\times10^{-4}$ weight decay. The total training epoch is set to 240 and the learning rate is divided by 10 at 150th, 180th and 210th epochs. The initial learning rate is set to 0.01 for MobileNet/ShuffleNet-series architecture and 0.05 for other architectures. The mini-batch size is set to 64.

Table~\ref{Table:app_seq} gives additional results with different initial learning rates. It is shown that the student accuracy always stays at about 50\% when the learning rate ranges from 0.01 to 0.5, which indicates the difficulty of training a satisfactory student classifier from scratch. In contrast, our SimKD achieves \textbf{78.08 $\pm$ 0.15} test accuracy without any classifier retraining but just reusing the pre-trained teacher classifier. 

\subsection{Comparison of Loss Function}
The default feature alignment loss in our main submission is implemented in the preceding layer of the teacher classifier with a $\ell_2$ loss. Its result is reported in the second column of Table~\ref{Table:app_loss}. 
Another implementation is to calculate the loss in the succeeding layer of the teacher classifier with a loss function $\|\boldsymbol{W}^{t}\boldsymbol{f}^{t}-\boldsymbol{W}^{t}\mathcal{P}(\boldsymbol{f}^{s})\|^2_2$, and we report its results in the third column of Table~\ref{Table:app_loss}.
From Table~\ref{Table:app_loss}, we find that our default feature alignment loss performs best. Moreover, the gradient comparison of different loss functions indicates that the effect of ``Output ($\ell_2$)'' is to calibrate the gradient of  ``Input ($\ell_2$)'' with a symmetric matrix ${\mathbf{W}^t}^\mathrm{T}\mathbf{W}^t$.


\subsection{Comparison of Pruning Ratio}
Table~\ref{Table:app_pruning-1} and~\ref{Table:app_pruning-2} present the top-1 test accuracy and the cost of pruning ratio (the first element in parenthesis) of SimKD versus different dimension reduction factors. We also provide the ratio of the projector parameters to the student parameters (the second element in parenthesis) for comparison.
We make those results bold when SimKD achieves state-of-the-art performance and the added projector only requires less than or about 3\% pruning ratio cost. 

In some cases such as ``MobileNetV2x2 \& ResNet-32x4'' and ``ShuffleNetV1 \& ResNet-32x4'' with $r=8$, we can see that the pruning ratios of SimKD are even higher than the vanilla KD training, and all competitors accordingly. Moreover, SimKD achieves the second best performance on ``ShuffleNetV2 \& ResNet-32x4'' with $r=8$ (SimKD: 77.49\%, the best performance is achieved by SemCKD:  77.62\%), 
``ShuffleNetV2x1.5 \& ResNet-32x4'' with $r=8$ (SimKD: 78.96\%, the best performance is achieved by SemCKD:  79.13\%), and ``ShuffleNetV2 \& ResNet-110x2'' with $r=4$, (SimKD: 77.35\%, the best performance is achieved by SemCKD: 77.67\%). Although the projector needs retaining during the whole training and test stages, a series of trade-off experiments between test accuracy and pruning ratio show that the extra parameters it brought are negligible in most cases. 

We further extend our technique to the situation where more deep teacher layers are reused for student inference and analyze the accompanying trade-off between accuracy enhancement and complexity increase.
As shown in Table~\ref{Table:app_reuse}, ``SimKD+'' and ``SimKD++'' achieve higher performance than ``SimKD'' but they also bring about a sharp drop of the pruning ratio, which indicates that simply reusing the final teacher classifier strikes a good balance between performance and parameter complexity.
\begin{table}[t]
	\centering
	\begin{tabular}{l|cc}
		\toprule
		& Test Accuracy & Pruning Ratio\\
		\midrule
		Student & 73.09 $\pm$ 0.30 & 83.40\%\\
		SimKD 	& \textbf{78.08 $\pm$ 0.15} & \textbf{80.52\%} \\
		SimKD+ & \textbf{78.47 $\pm$ 0.08} & 19.21\%  \\
		SimKD++ & \textbf{78.88 $\pm$ 0.05} & 15.97\% \\
		\midrule
		Teacher & 79.42 & 0\% \\
		\bottomrule
	\end{tabular}
	\caption{Comparison of reusing different teacher layers.}
	\label{Table:app_reuse}
\end{table}

\subsection{Visualization}
We adopt ResNet-8x4 as the student model and ResNet-32x4 as the teacher model for visualization experiments. 

\begin{figure}[h]
	\centering
	\begin{subfigure}{0.48\columnwidth}
		\includegraphics[width=0.98\columnwidth]{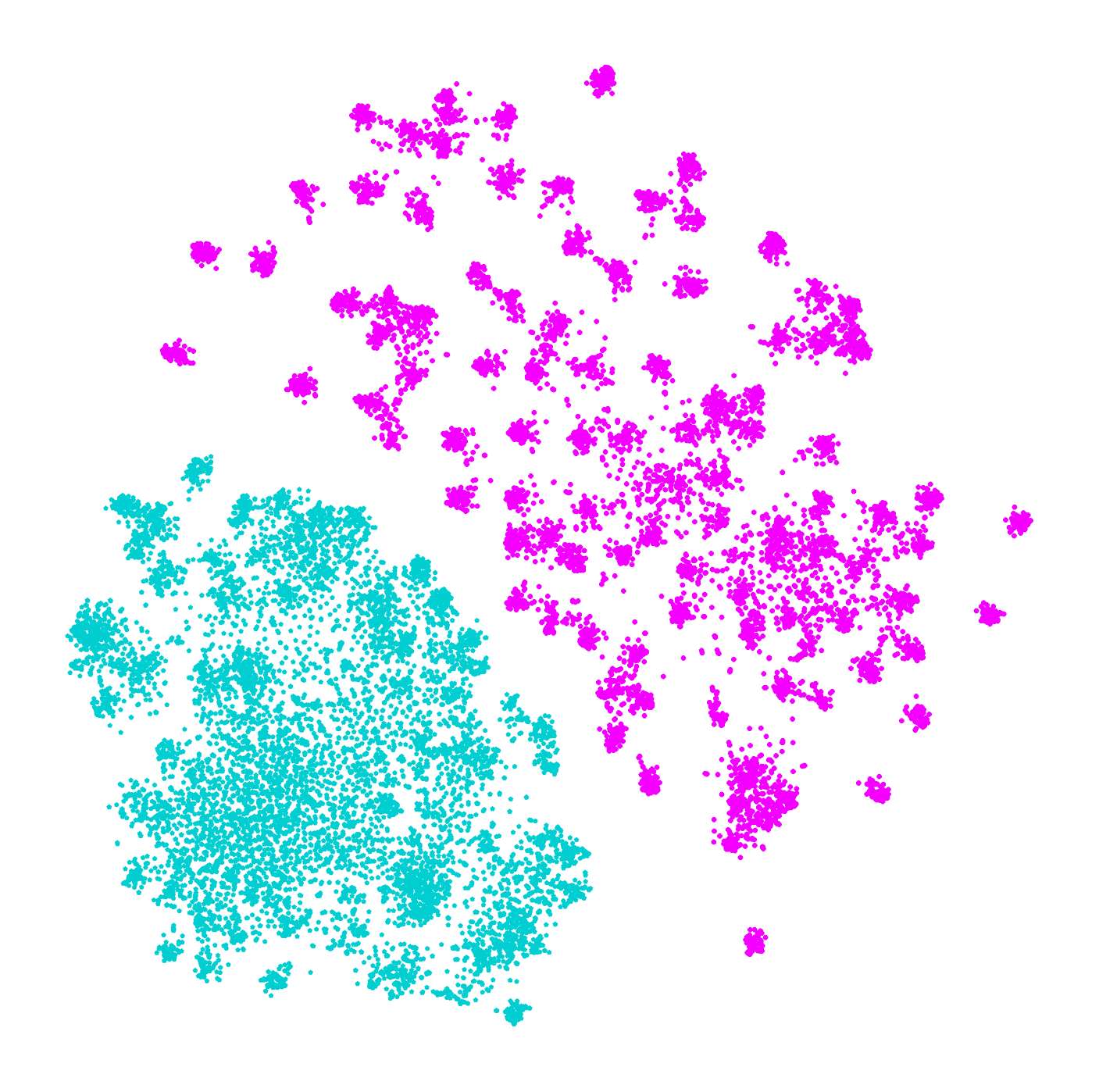}
		\caption{Vanilla KD \cite{hinton2015distilling}.}
		\label{fig:vis-a}
	\end{subfigure}
	\hfill
	\begin{subfigure}{0.48\columnwidth}
		\includegraphics[width=0.98\columnwidth]{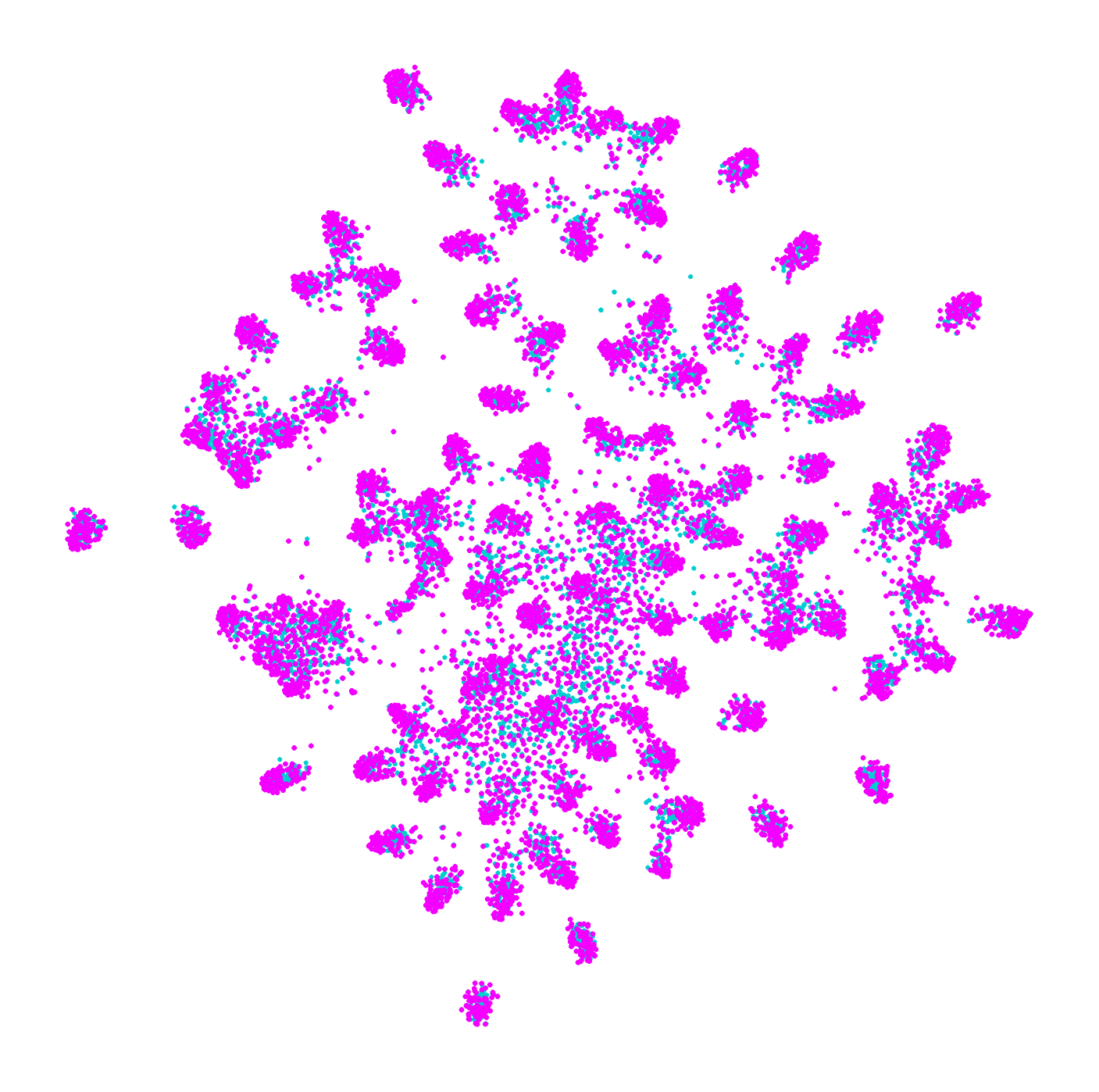}
		\caption{Our SimKD.}
		\label{fig:vis-b}
	\end{subfigure}
	\caption{Visualizations of all test images from CIFAR-100 with t-SNE \cite{vandermaaten2008visual}. Features extracted by the teacher and student models are depicted with magenta and cyan colors, respectively, and they are almost indistinguishable in our SimKD. Best viewed in color.}
	\label{fig:app_vis_100}
\end{figure}

Ten randomly selected classes in the main submission includes ``road'', ``bee'' , ``lawn\_mower'', ``bottle'', ``shrew'', ``bridge'', ``man'', ``mouse'', ``sweet\_pepper'' and ``cattle''. 
We further visualize all 100 classes on CIFAR-100 with t-SNE in Figure~\ref{fig:app_vis_100}. The visualization results show that with the help of a simple $\ell_2$ loss, the extracted features from teacher and student models become almost indistinguishable in SimKD, which ensures the student features to be correctly classified with the reused teacher classifier later. 

{\small
	\bibliographystyle{ieee_fullname}
	\bibliography{egbib}
}

\end{document}